\documentclass[a4paper]{IEEEtran} 

\usepackage{cite}
\usepackage[pdftex]{graphicx}
\usepackage{algorithm}
\usepackage{algpseudocode}
\usepackage{mdwmath}
\usepackage{mdwtab}
\usepackage{caption}
\usepackage[font=footnotesize]{subfig}
\usepackage{amssymb}
\usepackage{amsmath}
\usepackage{amsfonts}
\usepackage[utf8]{inputenc}
\usepackage{cleveref}
\usepackage{soul}
\usepackage{tabularx, booktabs}
\usepackage{multirow}
\usepackage{grffile} 
\usepackage{rotating}         
\usepackage{threeparttable}   

\DeclareGraphicsExtensions{.pdf,.jpeg,.png}

\usepackage{newfloat}
\usepackage{tikz}
\newcommand{\framedbox}[2][0.90\columnwidth]{     
  \tikzstyle{mybox} = [draw=black,line width=1pt,inner sep=8pt]
  \begin{tikzpicture}   
   \node [mybox] (box){\begin{minipage}%
   {#1}{#2}
   \end{minipage}};
  \end{tikzpicture}
}
\DeclareFloatingEnvironment[listname=Box,name=Box]{story}
\makeatletter
\newcommand*{\textoverline}[1]{$\overline{\hbox{#1}}\m@th$}
\makeatother

\def\eg{\emph{e.g., }}
\def\ie{\emph{i.e., }}
\def\wrt{\emph{w.r.t. }}

\title{Vehicle Ego-Lane Estimation with Sensor Failure Modeling}

\begin{document}


\DeclareRobustCommand*{\IEEEauthorrefmark}[1]{%
  \raisebox{0pt}[0pt][0pt]{\textsuperscript{\footnotesize\ensuremath{#1}}}}

\author{\IEEEauthorblockN{Augusto Luis Ballardini\IEEEauthorrefmark{1}, 
                      Daniele Cattaneo\IEEEauthorrefmark{3}, 
                      Rubén Izquierdo\IEEEauthorrefmark{1}, 
                      Ignacio Parra\IEEEauthorrefmark{1},\\
                      Andrea Piazzoni\IEEEauthorrefmark{4},
                      Miguel Angel Sotelo\IEEEauthorrefmark{1},
                      Domenico Giorgio Sorrenti\IEEEauthorrefmark{2},
                      }
                      
\IEEEauthorblockA{\IEEEauthorrefmark{1}Computer Engineering Department, Universidad de Alcala, Spain}
\IEEEauthorblockA{\IEEEauthorrefmark{2}Computer Science Department (DISCO), Università degli Studi di Milano-Bicocca, Italy}
\IEEEauthorblockA{\IEEEauthorrefmark{3}Department of Computer Science, University of Freiburg, Germany}
\IEEEauthorblockA{\IEEEauthorrefmark{4}Interdisciplinary Graduate School,
Nanyang Technological University, Singapore}
}

\maketitle

\begin{abstract}
We present a probabilistic ego-lane estimation algorithm for highway-like scenarios that is designed to increase the accuracy of the ego-lane estimate, which can be obtained relying only on a noisy line detector and tracker. The contribution relies on a Hidden Markov Model (HMM) with a transient failure model.
The proposed algorithm exploits the OpenStreetMap (or other cartographic service) road property ``lane number'' as the expected number of lanes and leverages consecutive, possibly incomplete, observations.
The algorithm effectiveness is proven by employing different line detectors and showing we could achieve much more usable, \ie stable and reliable, ego-lane estimates over more than 100Km of highway scenarios, recorded both in Italy and Spain.
Moreover, as we could not find a suitable dataset for a quantitative comparison with other approaches, we collected datasets and manually annotated the Ground Truth about the vehicle ego-lane. Such datasets are made publicly available for usage from the scientific community.
\end{abstract}


\section{Introduction}
Autonomous vehicles require an accurate understanding of the surrounding environment in order to safely plan their actions. One such fundamental perception task concerns the localization of the vehicle. Autonomous vehicles cannot always rely on global positioning based on Global Navigation Satellite System (GNSS) signals (\eg GPS, Beidu, Glonass, Galileo, etc.) because signals from satellites use to undergo multi-paths and physical barriers, leading sporadically to very poor position accuracy or even to no estimate at all.
Therefore navigation modules usually couple the GNSS systems with cartographic maps and methods that leverage the road network graph as well as other common features \cite{Steder2015, Raaijmakers2015, Gevers2014, Ballardini2016, Ballardini2017}, like buildings, crossings or roundabouts, which are retrieved from cartographic services like OpenStreetMap. Therefore maps represent an important piece of information to exploit as prior in vehicle localization.
Even though methodologies based on GNSS and maps, usually known as \emph{lock-on-road} procedures, see \eg \cite{Alonso2012,Ballardini2015}, lead to remarkable increases in the localization accuracy, they still do not use to achieve lane-level localization, \ie accuracies in the order of 0.1m \cite{Levinson2011}.

Lane-level localization is actually a term with broad meaning and it usually might refer to two different problems: on the one hand it might refer to the determination of the lane currently occupied by the vehicle, a problem which is also known as \emph{host-lane} or \emph{ego-lane} estimation; on the other hand it might refer to the estimation of the lateral position of the vehicle inside the lane or whole road. The latter problem is relevant for the lower level control of the vehicle, while the first is relevant in the higher-level (tactical) control of the vehicle, \ie trajectory and manoeuvre planning. Solutions to the latter live in $\mathbb{R}$, while solutions to the first live in $\mathbb{N}$. This work deals with the ego-lane estimation problem.

In both cases, some form of optical detection of the road lines is usually exploited. Quite frequently images obtained from the forward-looking camera(s) of the vehicle are used, and processed in order to detect the road lines. Typically, such lines are in turn processed to infer the vehicle ego-lane at the time of the image capture.

In this paper we present a probabilistic algorithm aimed at enhancing the ego-lane estimation obtained from a line detector. The algorithm exploits a GNSS measure as well as the number of lanes of the road, retrieved from a service like OpenStreetMap, as prior. Differently from other works available in the literature, ours is a modular, hence reusable, algorithm for improving the ego-lane estimation that could be obtained from a generic line detector. Our algorithm relies on a Hidden Markov Model (HMM) with a transient failure model, which allows us to accommodate inaccurate or missing road line detections.

The paper is organized as follows. Section~\ref{sec:related-work} provides a short overview of the existing ego-lane estimation literature, Section~\ref{sec:proposed-algorithm} describes the proposed algorithm and Section \ref{sec:experimental-section} introduces the experimental configurations and datasets. Finally, Section~\ref{sec:discussion} critically presents the experimental results of the algorithm, and is followed by concluding remarks.

\section{Related work}\label{sec:related-work}
Ego-lane estimation for autonomous driving has been extensively investigated in the last decades. The first achievements were obtained by the group of Prof. Dickmanns \cite{Dickmanns1992}, they introduced a road representation model based on clothoids, which were then updated with the image measurements by using Kalman filters.

Basing on these results, an active research has been conducted in the subsequent years \cite{Pomerleau1995, Bertozzi1998, Taylor1996, Wang2004a}. Heterogeneous modeling techniques for the lane markings, including parabolas, clothoids, poly-lines or b-splines, were proposed, typically computed from images, after some preprocessing phases designed to remove clutter and irrelevant areas.

One of the most difficult tasks that are to be solved to identify the ego-lane is the detection of the road surface. Achieving a good discrimination of the road surface from other parts is crucial since it is the basis for further processing, but this detection is usually adversely affected by the large amount of clutter usually found on real roads.
While faded road markings, unusual or specific weather conditions, or even light variations might severely affect the road surface detection, the visibility of the road surface is quite frequently hampered by the presence of other vehicles, thus requiring different considerations to solve the problem.

Most of the current Advanced Driver Assistance Systems (ADAS), like Lane Departure Warning (LDW) or Adaptive Cruise Control (ACC), require just a partial understanding of the whole observed scene, like the lines of the vehicle's lane or the lane crossing points, in highway-like scenarios \cite{BarHillel2014,Mccall2006a}.

For what concerns sensing, even though LIDAR-based algorithms sport the advantage of active lightening, vision-based algorithms are, as for today, the most frequently used sensing approach for line detection and ego-lane estimation, since road markings are designed to be human-visible in mostly all driving conditions, see \eg \cite{BarHillel2014}, which has an excellent review of approaches in lane localization.

With the objective to pursue lane-level localization, the authors in \cite{stefanorf2016} propose to exploit the objects present in the surrounding of the vehicle and to describe the probabilistic dependencies between the object measurements, by means of a factor graph model. A similar proposal comes from the authors of \cite{Flade2016}, where Histogram of Oriented Gradients are used to align the images acquired from a front facing camera to the road lane markings, to improve the vehicle localization.

Many authors propose, in order to increase the performance of ego-lane estimation algorithms, to exploit additional road information gathered by map services as well as information provided by GNSS. In this regard, an interesting approach is presented in \cite{Gao2009}, where the authors tackled the ego-lane estimation as a scene-classification problem. They infer the lane number in a holistic fashion, leveraging both spatial information and objects around the vehicle, and finally training the best classifier with different learning algorithms. In \cite{Kim2008} the author presented a robust lane-detection-and-tracking algorithm combining a particle filtering technique for lane tracking and RANSAC for the detection of lane boundaries. The work detects left and right lane boundaries separately, without exploiting fixed width lane models, and combining lane detection and tracking within a common probabilistic framework.

The authors in \cite{Kuhnl2013,Rabe2016}, respectively in highway and urban scenarios, propose to exploit boosting classifiers and particle filtering approaches. A similar research was performed by \cite{Lee2015}, where multiple evidence from a visual processing pipeline was combined within a Bayesian Network approach. 

Closer to our proposal are the works in \cite{Nieto2008,Jiang2010,Kang2014}, where the authors specifically address the multiple-lane detection problem. In \cite{Nieto2008} multiple lane detections are performed after a first processing phase, where the authors identify the ego-lane geometry. Then, adjacent lanes are first hypothesized and then tested, assuming same curvature and width for all lanes, a fair assumption for most of multi-lane roads, including highways. Similarly, the work proposed in \cite{Jiang2010} also considers highway scenarios and parallel lane markings, with respect to the detected ego-lane. More recently, the authors in \cite{Kang2014} proposed a multi-lane detection algorithm based also on a hypothesis generation and testing scheme, ensuring an accurate geometric estimation by means of a robust line fitting pipeline and vanishing point estimation.


Differently from the other contributions, where the authors propose new detection pipelines for the ego-lane estimation problem, here we introduce a generic scheme aimed at improving the ego-lane estimation capabilities of potentially every line detector. Also, the output of a lane detection algorithm could be fed into our algorithm, to increase its performance in ego-lane estimation. Our aim was to enhance the localization capabilities of the scene understanding framework proposed in~\cite{Ballardini2015}, introducing a lane-awareness module capable of reducing localization errors in highway-like environments. As a by-product of the modularity of the framework, we could easily compare the localization results obtained with and without the new proposed algorithm.

\section{Proposed algorithm}\label{sec:proposed-algorithm}
The goal of the proposed algorithm is to estimate the vehicle ego-lane in highway scenarios, \ie when the topology of the roadway does not change, \eg because of exit ramps, bifurcations, etc. The input of the proposed algorithm are both a global localization, at the level of accuracy provided by today GNSS, and the detections of the road line markings.

The algorithm is designed to tolerate occasional temporary failures of the underlying line detector as well as its noisy measurements. A line detector is a software component that detects and tracks the relative position of both dashed and continuous road lines, with respect to the vehicle.

Our algorithm relies on a probabilistic model of the functioning of the ego-lane estimation, and is designed to be independent from the line detector, so we can compare the results of our algorithm when working on the output of different line detectors.

\begin{figure}
  \centering
    \includegraphics[width=1.00\columnwidth]{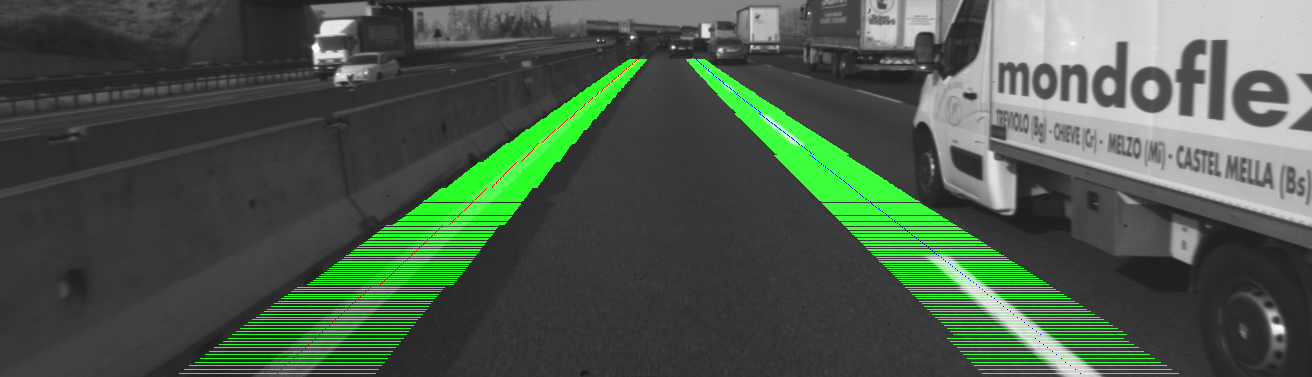}
    \caption{An example of a moderately congested condition on the A4 (Turin - Milan - Venice - Trieste) highway, Italy. Even at this moderate level of congestion most road markings are hidden by traffic.}
  \label{fig:cluttered}
\end{figure}

\begin{figure}
  \centering
    \subfloat[]{\includegraphics[width=0.49\columnwidth]{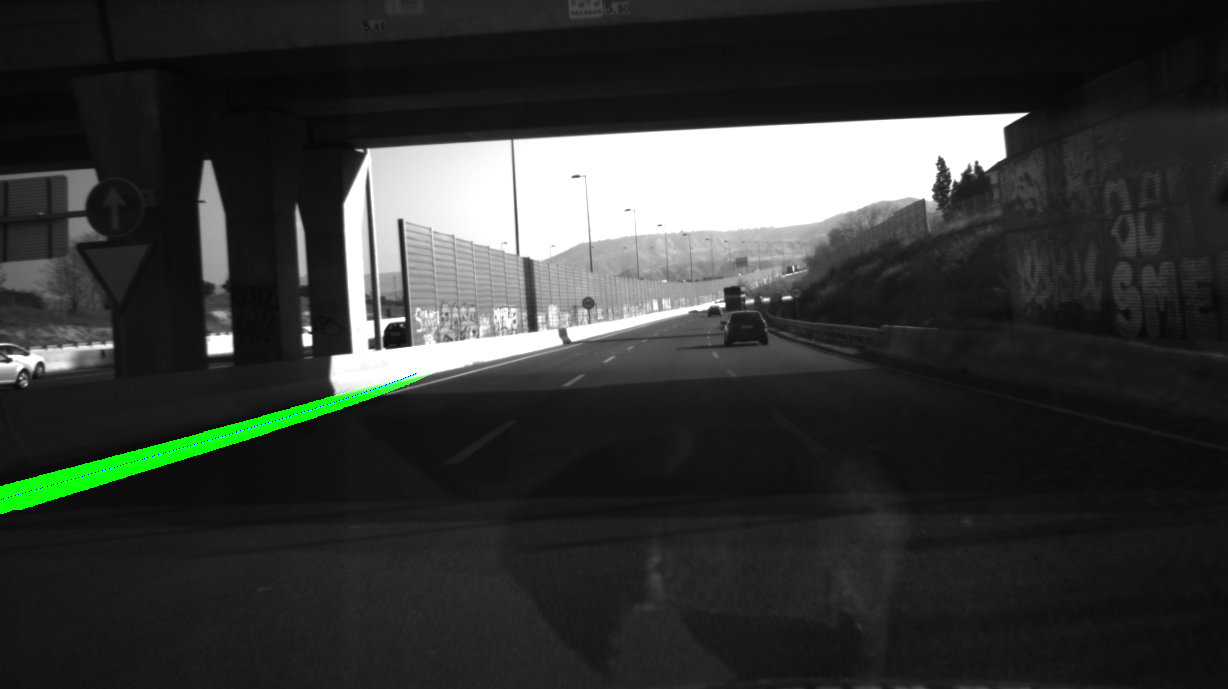}\label{fig:unasolariga}}\hspace{\fill}
    \subfloat[]{\includegraphics[width=0.49\columnwidth]{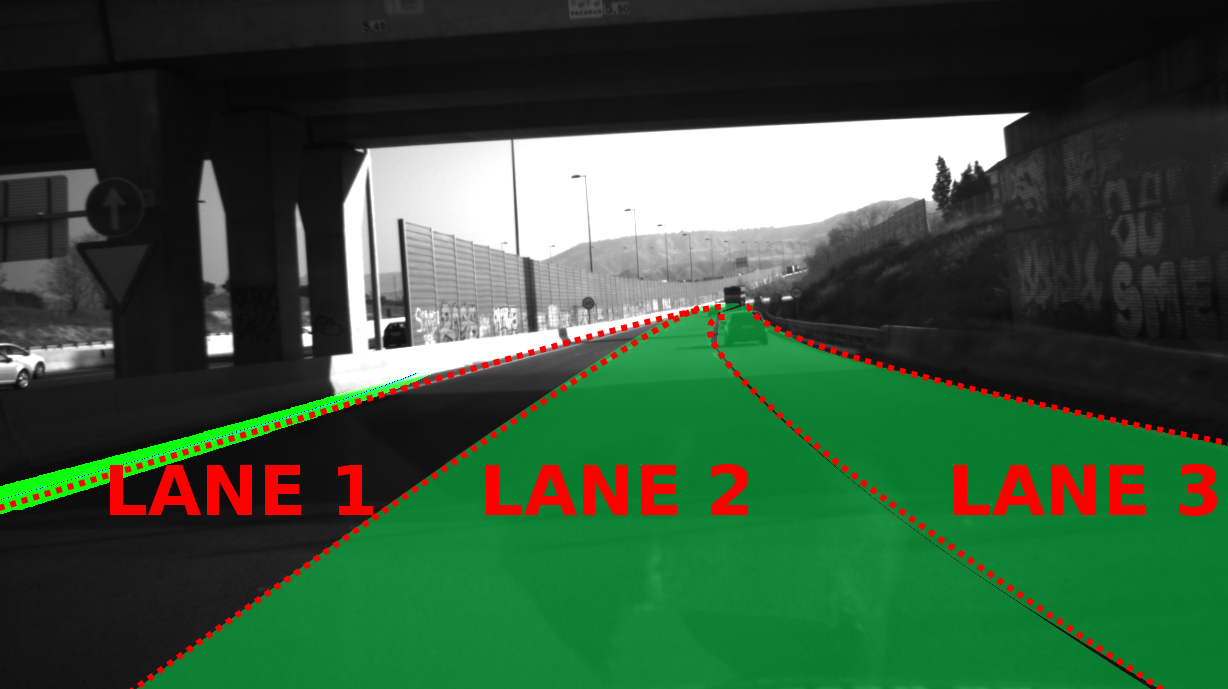}\label{fig:unasolariga00}}
    \caption{In (a) only one line out of four is detected, thus the highlighted lanes in (b) have a higher probability of being the vehicle ego-lane, as implicated by the distances to the detected lines.}
  \label{fig:unasolariga-multipla}
\end{figure}

\begin{figure}
  \centering   
    \includegraphics[width=1.00\columnwidth]{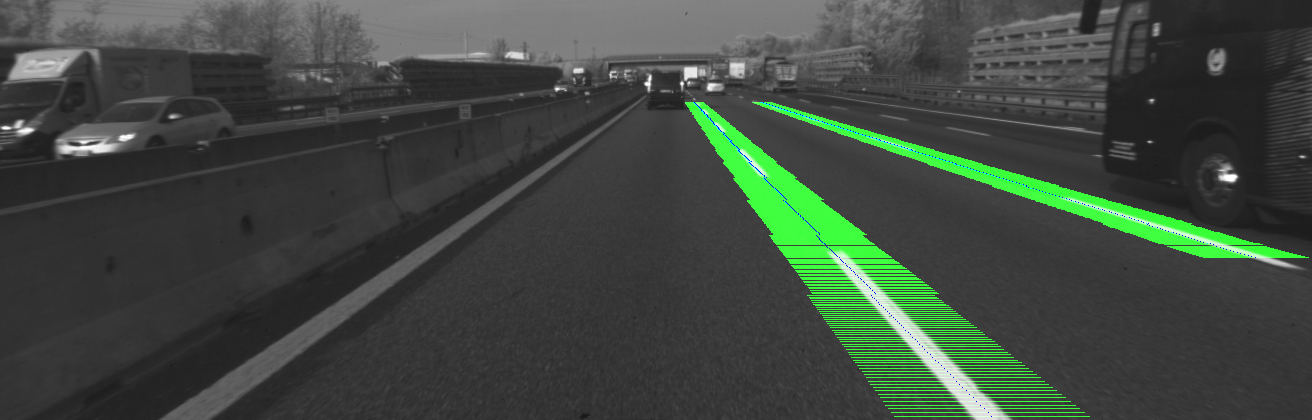}
    \caption{In this figure, just two out of the five lines are correctly detected and tracked. The shadow created by the Jersey barrier prevents the correct detection of the leftmost line, although in our opinion the line might be detected, perhaps with a different set of parameters, which might in turn bring in other misdetections. An error might also arise with dashed lines, whenever the space between two consecutive detected dashes is increased by some vehicle. The two rightmost lines are not detected because of their limited thickness; also in this case a different processing pipeline could detect such lines, usually introducing other errors.}
  \label{fig:onscreen-results}
\end{figure}

Indeed, the estimation of the vehicle ego-lane can be regarded as a consequence of the outcome of the line detection procedure. Actually, the position of all the road lines \wrt the vehicle, allows to determine the ego-lane by using simple geometric considerations, on a per-frame basis. Unfortunately, line detections are usually not fully reliable, being hampered by faded road markings, cluttering elements from the nearby traffic, or weather conditions, see \eg Figure~\ref{fig:cluttered}, \ref{fig:unasolariga} and \ref{fig:onscreen-results}.

However, we combine the line detections with an index about how reliable each detection is, we call this index \emph{Line Reliability Index} (LRI). This index, together with our proposed probabilistic model, allows an appropriate handling of the noisy output of the line detectors.

Furthermore, consider the situation depicted in Figure~\ref{fig:unasolariga}, surely a critical situation for ego-lane estimation. 
Even though the exact lane cannot be estimated from the only detected line, the distance measured from such line would allow to limit the uncertainty only to the compatible lanes, as depicted by the green highlighted lanes in Figure~\ref{fig:unasolariga00}.

\begin{figure}
  \centering   
    \includegraphics[width=1.00\columnwidth]{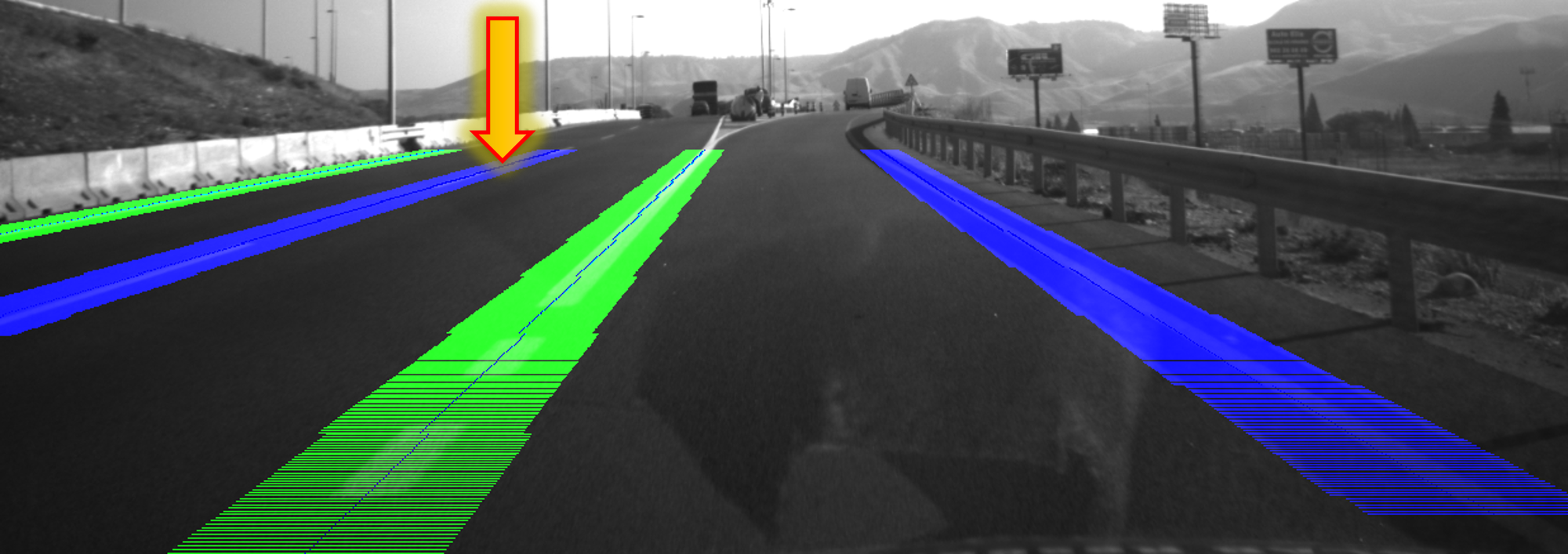}      
    \caption{If only the line indicated with the arrow were detected, the probability of being in Lane${\{1|2|3\}}$ would be \{0,\:0.5,\:0.5\}. The idea of exploiting the plausibility given by each line, repeated for all the detected lines, and together with our model, allows to solve the ego-lane estimation problem. Here the green and blue visually suggest the reliability of the lines (green is higher).}
  \label{fig:teaser}
\end{figure}

Our proposal is to tackle the ego-lane estimation with a probabilistic model, in order to allow the system to infer the ego-lane by leveraging consecutive, yet incomplete, observations over time. An HMM is proposed, with a Lane variable that can take $n$ values, corresponding to the number of lanes retrieved from an OpenStreetMap-like service.

\subsection{Line detection and tracking}\label{section:simple-line-detector}
In this section, we shortly describe the line detection and tracking algorithms used in the experimental activity. We first introduce a ``simple'' line detector and tracker that is capable to work with both a stereo and a monocular camera configuration, requiring the projection of the camera(s) to be calibrated \wrt the vehicle frame before its usage. This algorithm consists of the following steps:
\begin{itemize}
\item The contours of the road markings are extracted from the Bird Eye / Inverse Perspective view (BEV / IPV) of the left camera image and discarded if their area is below a threshold. To compute the BEV / IPV image an homography matrix is computed, basing on the intrinsic values of the projection model, and the extrinsic values \wrt the road surface. The contours in the BEV image are then determined using the algorithm proposed in \cite{suzuki1985topological}.
\item The algorithm then tries to fit, onto the detected contours, a fixed number of lines or clothoids (both the number and the type are parameters), trying to cover the highest number of contour areas; if the stereo configuration is available, the algorithm exploits it to exclude lines / clothoids not lying on the ground plane. The ground plane equation is evaluated using the output of the SGBM~\cite{sgbm} or the ELAS~\cite{Geiger2010ACCV} stereo-matching algorithm (this choice is a parameter). Concluding, both a monocular and stereo version of the algorithm is available.
\item The parameters of each line / clothoid are then updated by means of a Kalman filter.
\end{itemize}

W.r.t. the previous $k$ (\eg $k=10$) frames, the number of times the line is detected is taken as the \emph{Line Reliability Index} (LRI). Furthermore, again for each line, the LRI allows to set a flag, named \emph{isValid}, once the counter reaches its maximum value; to reset the flag the counting goes on with hysteresis, so that the flag is not reset until LRI goes below a certain fraction of its maximum value.

\begin{story}
  \caption{The line detector output for the image in Figure~\ref{fig:teaser}. The isValid flag is set to TRUE when LRI=10, and reset using a hysteresis counting procedure. A negative offset refers to lines on the left of the vehicle.}
  \label{box:box1}
  \framedbox[0.93\columnwidth]
  {  
  Line1: isValid = 1; continuous=1; LRI: 10; offset: -9.15m\\
  Line2: isValid = 0; continuous=0; LRI: 09; offset: -6.47m\\
  Line3: isValid = 1; continuous=0; LRI: 07; offset: -2.15m\\
  Line4: isValid = 0; continuous=1; LRI: 00; offset: +0.99m}
\end{story}

This simple line detector and tracker achieves good performances only under optimal illumination conditions and, as depicted in Figure~\ref{fig:teaser} and shown in the corresponding results in Box~\ref{box:box1}, dashed lines and shadows are not always handled correctly. However, it allowed to evaluate the effectiveness of our contribution, which is designed to enhance the vehicle ego-lane estimation by exploiting a noisy sensor as well as the \emph{road lane} properties gathered from an OpenStreetMap-like service.

Moreover, to effectively assess the proposed probabilistic model, we performed an extensive search to find other line detectors, sporting a measure of the reliability of the detection of each line. Although the literature about line generic detection is huge, as the problem has been investigated since the beginning of digital image processing, solutions dedicated to detecting road lines that are also available with software reduce, to the best of our knowledge, to the following two.
\begin{itemize}
\item The proposal by Mohamed Aly~\cite{aly2008real}, which exploits a robust approach based on a line / bezier line tracker and a RANSAC procedure. As far as their publicly available results show, this algorithm is able to detect all the lines on the road surface, making it one of the most complete and well performing algorithms. Nevertheless, on one hand the software that is publicly available does not include the tracking module, and on the other the number of parameters is overwhelming (about one hundred). These aspects make this software extremely hard to use. We have not been able to find a reasonably good configuration in the limited time available. In conclusion, this option, although very appealing, could not be used in our experimental activity.
\item The solutions by Hur~\cite{hur2013multilane} (MLD in the following), which detects a maximum of four lines, corresponding to the markings of the current lane as well as of the two neighboring lanes. This software could be adapted to our requirements for this work, \eg introducing a procedure for determining whether a line is dashed or continuous.
\end{itemize}
Neither of the two solutions can exploit a stereo camera configuration, and both rely on the extrinsic projection parameters to determine the distance of each line from the vehicle, exploiting the BEV / IPM image.

\subsection{Tentative vector and reliability of the whole detection}
To exploit the measurements provided by a line detector and tracker like the ones mentioned before, we derived a probabilistic (inverse) sensor model, which exploits both the spatial information carried by the lines and the LRIs, both produced by the line detector and tracker.
The processing pipeline is therefore composed as follows. First, the lines are sorted, in ascending order, based on their lateral offset \wrt the vehicle.
Then, a vector of counters, of the size of the number of lanes, is created. This vector is called \emph{tentative} (implied: \emph{distribution of the belief on the state as from the measurements}, as usual for an inverse sensor model). The values in this  vector are determined by iterating the following steps for all the valid lines, \ie the lines whose \emph{isValid} flag is set, taking into consideration whether each line is dashed or continuous.
\begin{itemize}
\item 1 is added to the i-th tentative vector value, if it is in accordance with the measurement, \ie if being in the i-th lane is compatible with the line position; this has the objective to cumulate the plausibility of being in the i-th lane, given the detected lines.
\item If the line has the continuous flag set, an additional \emph{Bonus Value} (BV) is added to the tentative vector position (based on the distance \wrt the line); this has the objective to represent the fact that continuous lines are more informative, as they usually are the leftmost / rightmost lines of the road.
\end{itemize}
After the evaluation of the line indicated with an arrow in Figure~\ref{fig:teaser}, the resulting tentative vector would be $[0;1;1]$.

During the iteration on all the lines, we also accumulate all the LRI counters, and compute the fraction \wrt the maximum LRI value times the current number of expected lines. This value is taken as an index of the overall reliability of all the line detections of the current frame, namely a Whole Output Reliability (WOR) of the detector. This in turn can be taken as an observation of the sensor being properly functioning or not.
	
It has to be noticed that some of these rules could be not adequate for all line detectors. For example, if a line detector could not provide a \emph{continuous} flag or a reliability index for each line, the set of rules must be modified, in order to provide a frame-level tentative vector and an overall WOR.

\subsection{HMM with Transient Failure Model}\label{hmm-with-transient-failure-model} 
To tackle the unavoidable problem of sensor failures, \ie the cases of the sensor not being able to correctly determine the value of the ego-lane state variable, we applied a filtering algorithm based on an HMM, which includes a state representing the functioning of the sensor itself. For an introduction to HMM see \cite{russellnorvig}. The proposed model allows to take advantage of incomplete and/or noisy road line observations in a probabilistic fashion, to better estimate the current ego-lane as well as whether the sensor is properly working or not. In our opinion, this extra degree of freedom, \ie the explicit modeling of the sensor functioning state, allows a better performance, \wrt considering the ego-lane as the only unknown, as it gives an extra area of accommodation for matching the unknown value of the state variables to the observations.

\begin{figure}
	\centering   
	\includegraphics[width=1.00\columnwidth]{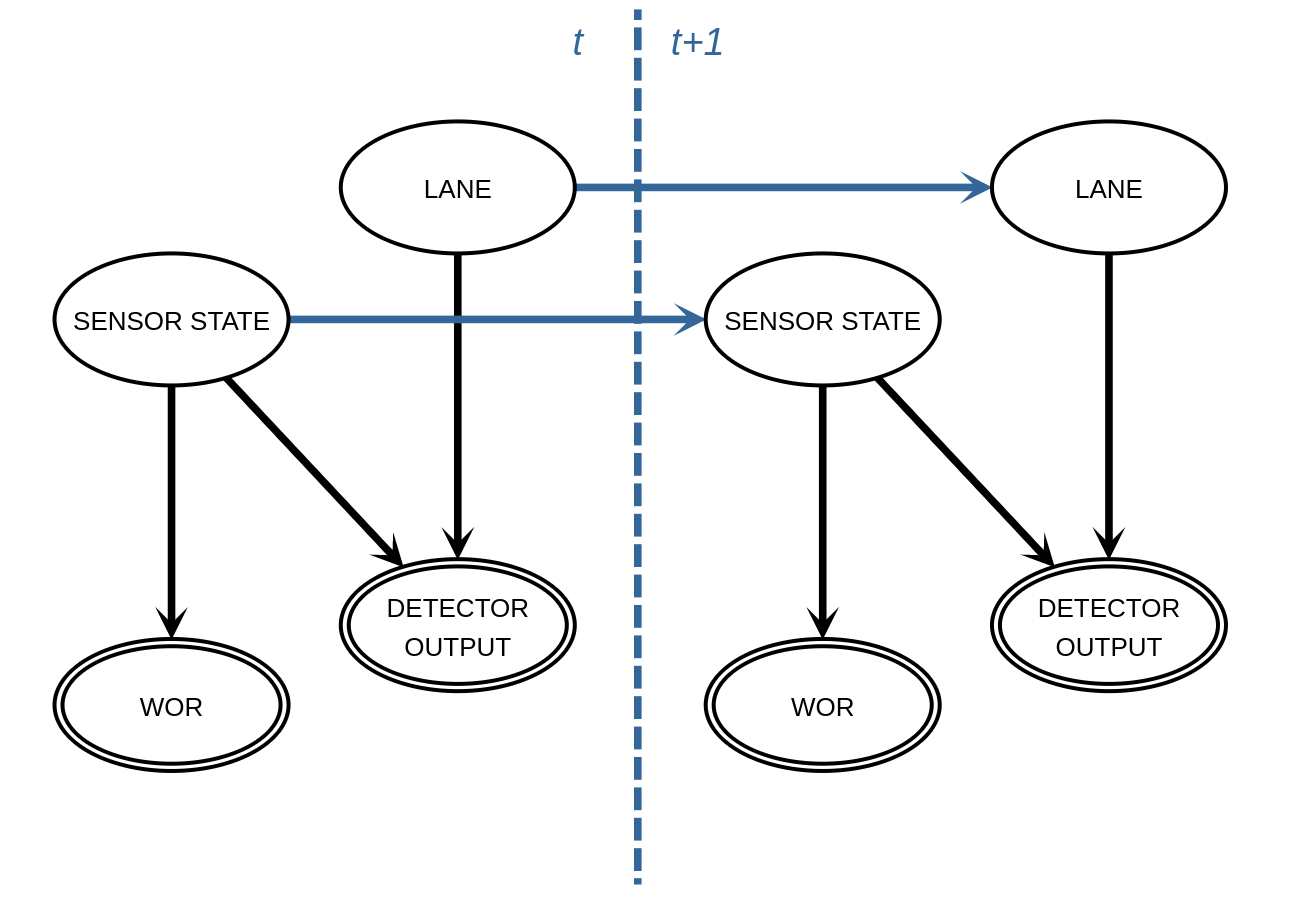}
	\caption{Two time frames of the HMM. The single circled variables are hidden, and the double circled variables are observable.}
	\label{fig:model}
\end{figure}

The HMM implements a filtering procedure over discrete random variables, where each iteration depends on the parameterization in Equation~\ref{eq:HMM}, see below for an explanation of the parameters.
\begin{equation}\label{eq:HMM}
\text{HMM}(n,\sigma_1,\sigma_2,p_1,p_2,p_3,p_4,BV)
\end{equation}
There are four variables, for each time frame, see Figure~\ref{fig:model}:
\begin{itemize}
\item Lane: represents the lane where the vehicle is (believed to be) located. This is a scalar variable, taking discrete values, which identify one of the road lanes.
\item Sensor State (also SS in the following): a scalar variable, taking one out of two values, which represents whether the sensor is properly working or not, and the taken values could be OK or BAD.
\item Detector Output: as we developed an inverse sensor model, this  variable is homogenous to the Lane variable; in principle it would be a scalar that can assume one out of $n$ values, in practice, to represent a belief over the ego-lane, it will be an $n$ components vector, summing to 1, where each one represents the probability of that lane being the vehicle ego-lane.
\item Reliability Index (WOR): analogously to Detector Output, here we have in principle a scalar variable, taking one value out of the ${OK, BAD}$ possible values; in practice WOR will be a 2 components vector, where each of the 2 values represents respectively how much the detector is believed to be properly working or not properly working.
\end{itemize}

The dependencies between these variables are described using Conditional Probability Tables (CPTs), \ie tables that describe the probability distribution of a variable (the values the variable can take are in the columns), given the state of its parents in the graph (the values the state variables can take are in the rows) of the model in Figure~\ref{fig:model}. In other words, the CPT tables allow to forward compute the expectations on the variables, which will be then updated with the observations.

The first two variables (Lane and Sensor State) have temporal dependencies, while the others only depend from variables within the same time frame.

Table~\ref{tab:cptlane} basically describes the dynamics of the lane variable, \ie the dynamics of the vehicle in the roadway. Thus, given a value of Lane at time $t$, we expect the Lane variable at time $t+1$ to follow a normal distribution with the Lane at time $t$ as mean and $\sigma^{2}_1$ as variance. This description needs a refinement, which is below; its functioning is indicated with $\mathcal{F}$, applied to a normal. The variance $\sigma^{2}_1$ controls how likely is for the vehicle to change lane during a time interval. Notice that $\sigma^{2}_1$ is set according to both the actual speed of a vehicle in changing lane and the frequency of the ego-lane estimation, \ie the image frame rate. For the belief on the lane to have the discrete domain of lanes, we transform ($\mathcal{F}$) the normal into a discrete set of values as follows: consider the domain of the probability density function (PDF) to be orthogonal to the lane axis. The PDF is fixed on the center of the lane as in the Table~\ref{tab:cptlane}; the integral across the lane is taken, from one sideline to the other. The lane widths could be assumed known, which is the approach taken in this work or, if available, they could be retrieved from an an OpenStreetMap-like service, as for the number of lanes. As the numbers obtained integrating across the lanes does not sum up to 1, \eg because of the infinite extension of the normal vs. the finiteness of the roadway, the last step represented by $\mathcal{F}$ is their normalization.

\begin{table}[H]
	\centering
	\caption{Lane CPT}
	\begin{tabular}{c|c|c|c|c|}
		\cline{2-5}
		& \multicolumn{4}{c|}{Lane$_{t+1}$} \\ \hline
		\multicolumn{1}{|c|}{Lane$_t$} & 1 & 2 & \ldots & $n$ \\ \hline
		\multicolumn{1}{|c|}{1} & \multicolumn{4}{c|}{$\mathcal{F}(\mathcal{N}(1,\,\sigma^{2}_1))$} \\ \hline
		\multicolumn{1}{|c|}{2} & \multicolumn{4}{c|}{$\mathcal{F}(\mathcal{N}(2,\,\sigma^{2}_1))$} \\ \hline
		\multicolumn{1}{|c|}{ \ldots} & \multicolumn{4}{c|}{ \ldots} \\ \hline
		\multicolumn{1}{|c|}{$n$} & \multicolumn{4}{c|}{$\mathcal{F}(\mathcal{N}(n,\,\sigma^{2}_1))$} \\ \hline
	\end{tabular}
	\label{tab:cptlane}
\end{table}

Table~\ref{tab:cptsensor} describes the sensor dynamics \wrt its mistakes, which in turn is related to the conditions of the road markings, and to the lightening conditions. If the sensor is working properly, it will remain in the properly working state with probability $p_1$, and switch to giving a wrong output with probability $(1-p_1)$. If the sensor is not in a properly working state, then it will remain in such a state with probability $p_2$, and switch back to properly working with probability $(1-p_2)$. This idea, by suitably setting $p_1$ and $p_2$, aims to represent the real experience of the sensor working properly most of time, and then presenting a sequence of frames with the sensor providing a wrong output.

\begin{table}[H]
	\centering
	\caption{Sensor State CPT}
	\begin{tabular}{c|c|c|}
		\cline{2-3}
		& \multicolumn{2}{c|}{Sensor State$_{t+1}$} \\ \hline
		\multicolumn{1}{|c|}{Sensor State$_{t}$} & OK & BAD \\ \hline
		\multicolumn{1}{|c|}{OK} & $p_1$ & $1-p_1$ \\ \hline
		\multicolumn{1}{|c|}{BAD} & $1-p_2 $&$ p_2 $\\ \hline
	\end{tabular}
	
	\label{tab:cptsensor}
\end{table}

Table~\ref{tab:cptdetlane} describes the detector output \wrt the state of its parents in the HMM: Sensor State and Lane. Therefore, the table has as many rows as the combinations of the values that the two conditioning variables can take, while it has as many columns as the values that the conditioned variable can take. As we have an inverse sensor model, the output of the detector is homogeneous to the Lane variable and we will have $n$ columns. If Sensor State is OK, we expect the detector to provide an output represented by a discrete set of numbers, drawn from normal distributions like it was for the Lane variable, \ie expected value set onto the real lane, $\mathcal{F}$ function steps like for the Lane variable, but variance described by the parameter $\sigma^{2}_2$, to represent the accuracy of the sensor in determining the ego-lane when properly working; this is clearly a different value \wrt the lane-change dynamics of the vehicle. However, if Sensor State is BAD, the detector output will be independent of the real lane, thus the output will be uniformly distributed.

\begin{table}[H]
	\centering
	\caption{Detector Output CPT}
	\begin{tabular}{cccccc}
		\cline{3-6}
		& \multicolumn{1}{c|}{} & \multicolumn{4}{c|}{Detector Output} \\ \hline
		\multicolumn{1}{|c|}{Sensor State} & \multicolumn{1}{c|}{Lane} & \multicolumn{1}{c|}{1} & \multicolumn{1}{c|}{2} & \multicolumn{1}{c|}{\ldots} & \multicolumn{1}{c|}{$n$} \\ \hline
		\multicolumn{1}{|c|}{\multirow{4}{*}{OK}} & \multicolumn{1}{c|}{1} & \multicolumn{4}{c|}{$\mathcal{F}(\mathcal{N}(1,\,\sigma^{2}_2))$} \\ \cline{2-6} 
		\multicolumn{1}{|c|}{} & \multicolumn{1}{c|}{2} & \multicolumn{4}{c|}{$\mathcal{F}(\mathcal{N}(2,\,\sigma^{2}_2))$} \\ \cline{2-6} 
		\multicolumn{1}{|c|}{} & \multicolumn{1}{c|}{\ldots} & \multicolumn{4}{c|}{\ldots} \\ \cline{2-6} 
		\multicolumn{1}{|c|}{} & \multicolumn{1}{c|}{$n$} & \multicolumn{4}{c|}{$\mathcal{F}(\mathcal{N}(n,\,\sigma^{2}_2))$} \\ \hline
		\multicolumn{1}{|c|}{\multirow{4}{*}{BAD}} & \multicolumn{1}{c|}{1} & \multicolumn{4}{c|}{$\mathcal{U}(1,\,n)$} \\ \cline{2-6} 
		\multicolumn{1}{|c|}{} & \multicolumn{1}{c|}{2} & \multicolumn{4}{c|}{$\mathcal{U}(1,\,n)$} \\ \cline{2-6} 
		\multicolumn{1}{|c|}{} & \multicolumn{1}{c|}{\ldots} & \multicolumn{4}{c|}{\ldots} \\ \cline{2-6} 
		\multicolumn{1}{|c|}{} & \multicolumn{1}{c|}{$n$} & \multicolumn{4}{c|}{$\mathcal{U}(1,\,n)$} \\ \hline
	\end{tabular}
	\label{tab:cptdetlane}
\end{table}

We can similarly define the CPT for WOR, the expected overall reliability of the output of the sensor, \eg as in Table~\ref{tab:cptri}. As in our model the WOR has only one parent, Sensor State, only two rows are needed in the Table. Notice though that a different model could be considered, where WOR depends on the Lane also, which might be more appropriate for certain situations and / or line detectors. The parameters $p_3$ and $p_4$ represents the probability of a correct evaluation of the Sensor State when the Sensor State is, respectively, OK and BAD.

\begin{table}[H]
	\centering
	\caption{WOR CPT}
	\begin{tabular}{c|c|c|}
		\cline{2-3}
		& \multicolumn{2}{c|}{WOR} \\ \hline
			\multicolumn{1}{|c|}{Sensor State} & OK & BAD \\ \hline
		\multicolumn{1}{|c|}{OK} & $p_3$ & $1-p_3$ \\ \hline
		\multicolumn{1}{|c|}{BAD} & $1-p_4 $&$ p_4 $\\ \hline
	\end{tabular}
	\label{tab:cptri}
\end{table}

\subsection{Inference}\label{sec:counting-scheme}
To perform inference with our model means to compute the most probable lane, given the tentative vector, \ie the output of the (inverse) sensor for the Lane and the Sensor State, which are in turn based on the output of the line detector and tracker.

To compute the belief on the HMM state (\ie Lane and Sensor State) at time $t+1$, we start from the HMM state at time $t$. Firstly, leveraging Table~\ref{tab:cptlane} and Table~\ref{tab:cptsensor}, we compute the expectation at time $t+1$ on these variables. \emph{Sensor State} is shortened in \emph{SS}.

\begin{equation}
 P(\text{Lane}_{t+1}\ |\ \text{Lane}_{t}) = P(\text{Lane}_{t}) \cdot \text{Lane CPT}
\end{equation}

\begin{equation}
P(\text{SS}_{t+1}\ |\ \text{SS}_t) = P(\text{SS}_t) \cdot \text{SS CPT}
\end{equation}

Then, in order to incorporate the new evidence carried by the Tentative vector and the WOR index, the Bayes formula has to be applied, so obtaining the belief over the HMM state at time $t+1$. There are in general two ways of applying the Tentative vector and WOR index as evidence in the inference. The first way is to simply consider the most probable value of the Tentative vector and WOR index as an hard evidence for the belief on the state. The other way, instead, is to consider the Tentative vector and the WOR index as soft evidence \cite{bilmes2004virtual}. Since this second way allows a more complete representation of the evidences, we have only considered this approach, also because the additional computation required for the inference is irrelevant, given the reduced size of the model.

\begin{figure}
 	\centering   
 	\includegraphics[width=1.00\columnwidth]{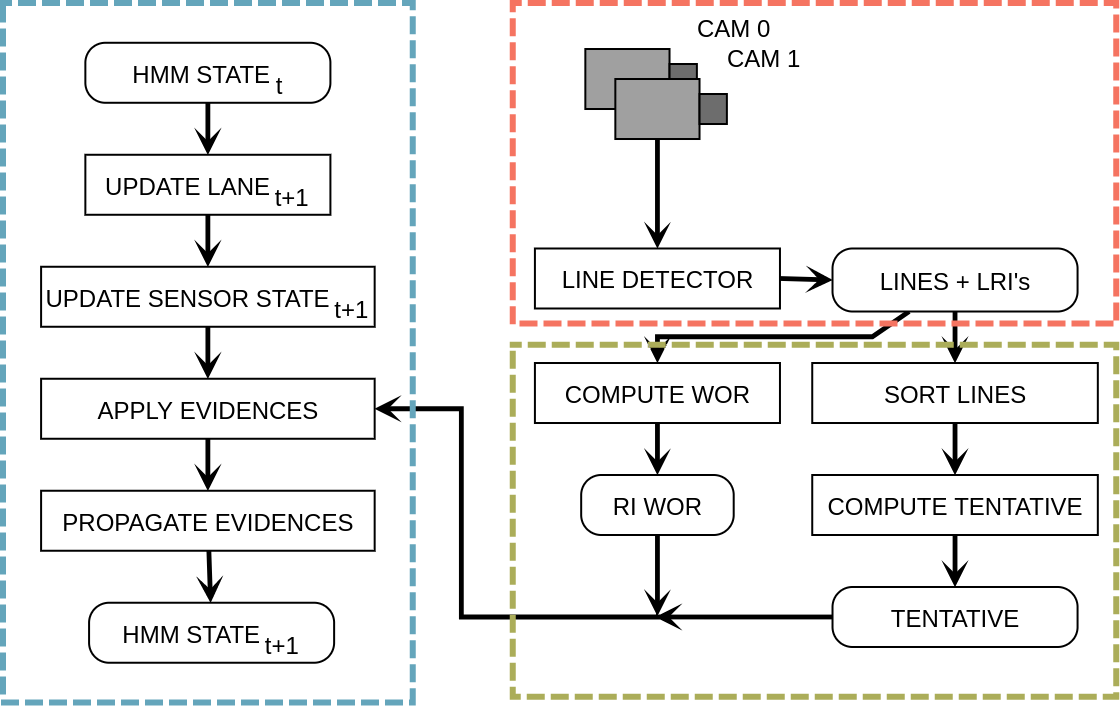}    
 	\caption{An overview of the proposed model.}
 	\label{fig:inference}
 \end{figure}

In Figure~\ref{fig:inference} is depicted an overview of the proposed model. 
The blue box is the core of our model, \ie the HMM, which is independent from the detector. The red box is the line detector and tracker, on which we tried to impose no constraint, keeping our model as general as possible. Lastly, the green box is the set of rules used to connect the line detector and tracker output to the HMM. This part may need to be changed with the output format of the line detector and tracker.
 
\section{Experimental Configuration}\label{sec:experimental-section}
\begin{figure}[!t]
  \centering   
  \subfloat{\includegraphics[width=1.0\columnwidth]{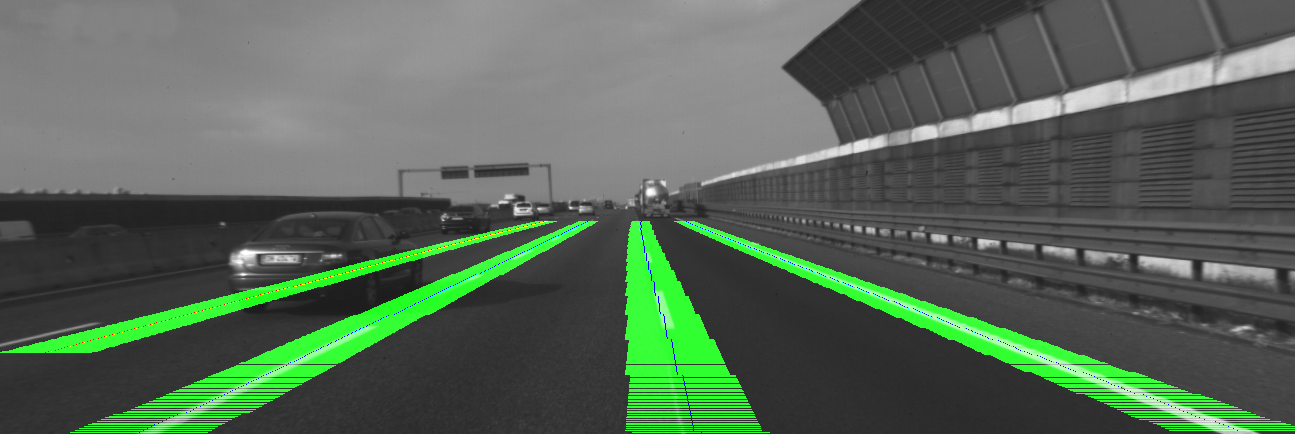}\label{fig:crossing-flag1}}\hspace{\fill}
  \smallskip   
  \subfloat{\includegraphics[width=1.0\columnwidth]{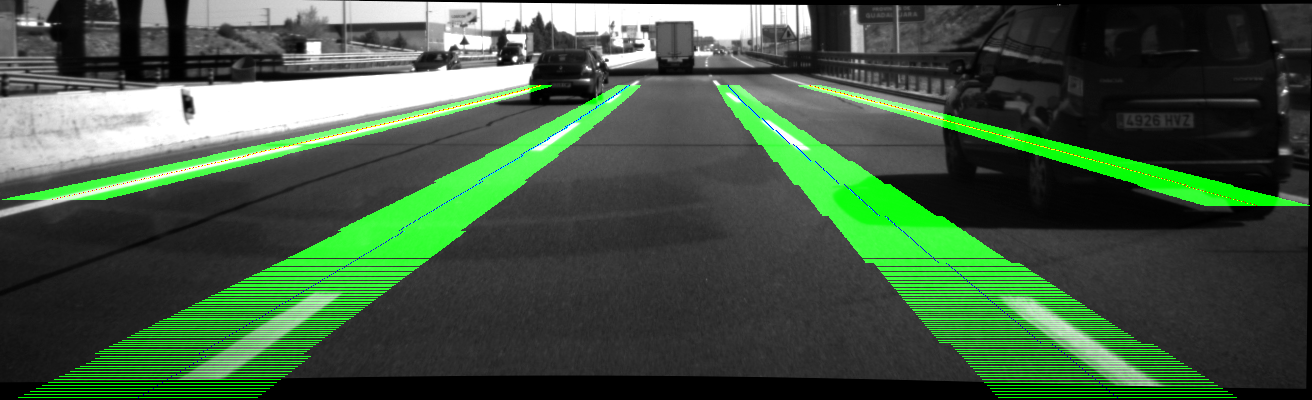}\label{fig:crossing-flag2}}\hspace{\fill}
    \caption{Two frames from the proposed annotated dataset, the detector output is  overlaid. In the top image, the vehicle was traveling in the A4 highway, Italy, and performing a lane change, \ie in the dataset the crossing flag is set. The bottom image depicts a frame from the A-2 highway, Spain.}
  \label{fig:crossing-flag}
\end{figure}

To effectively verify the improvements achieved by our model, we collected two datasets in real driving conditions. 
The first dataset was recorded in the A4 highway, Italy, from Bergamo to Milan. The second dataset is from the A2 highway area of Alcalá de Henares, Spain.
Both the datasets were recorded at 10 fps and have a resolution of 1312x540 and 1392x400 pixels respectively. 
Differently from standard datasets like KITTI, in which the highway sequences only contain few lanes, we drove our vehicles on wider highways with 3 and 4 lanes (Spain and Italy respectively), including more than 100 lane changes in the A4 highway sequences. 
We manually annotated the ground truth (GT) about the correct lane for more than 20K frames, considering as $Lane_1$ the leftmost lane as in~\ref{fig:unasolariga}. For each frame, we also included a ``crossing flag'' to indicate whether the vehicle is changing lane, so to exclude ambiguous lane assignments, see \Cref{fig:crossing-flag}.

For each experimental setting, the parameterization of~\Cref{eq:HMM} was empirically defined after an optimization phase, aimed at identifying the best parameter set \wrt the GT. Please, notice that the parameters are different since we changed either the detector configuration or the detector itself. The parameter values used during the experiments in Italy and Spain are reported in~\Cref{tab:full-parametrization-italy} and ~\Cref{tab:full-parametrization-spain}.

\begin{table}[]
\centering
\begin{threeparttable}
\caption{Parameterizations used for the different experimental settings - Italy ($n$ = 4).}
\label{tab:full-parametrization-italy}
\begin{tabular}{|c|c|c|c|c|c|c|c|c|}
\hline
Run  &\begin{tabular}[c]{@{}c@{}}Det.\\ Alg.\end{tabular}  & s1    & s2    & p1    & p2    & p3    & p4    & BV  \\ \hline                                              
01   &                           ML                        & 0.336 & 0.696 & 0.895 & 0.894 & 0.690 & 0.461 & 7   \\ \hline
02   &                           SL                        & 0.481 & 0.296 & 0.160 & 0.970 & 0.613 & 0.975 & 9   \\ \hline
03   &                           MC                        & 0.407 & 0.360 & 0.853 & 0.993 & 0.303 & 0.640 & 4   \\ \hline
04   &                           SC                        & 0.386 & 0.598 & 0.906 & 0.994 & 0.311 & 0.595 & 7   \\ \hline
05   &                           MLD                       & 0.324 & 0.707 & 0.223 & 0.963 & 0.779 & 0.873 & 1   \\ \hline
\end{tabular}
\begin{tablenotes}[para,flushleft]
M = mono, L = line, S = stereo, C = clothoid, MLD = \cite{hur2013multilane}
\end{tablenotes}
\end{threeparttable}
\end{table}

\begin{table}[]
\centering
\begin{threeparttable}
\caption{Parameterizations used for the different experimental settings - Spain ($n$ = 3.}
\label{tab:full-parametrization-spain}
\begin{tabular}{|c|c|c|c|c|c|c|c|c|}
\hline
Run   &\begin{tabular}[c]{@{}c@{}}Det.\\Alg.\end{tabular}  & s1    & s2    & p1    & p2    & p3    & p4    & BV\\ \hline                                              
06    &                           ML                       & 0.407 & 0.258 & 0.692 & 0.590 & 0.180 & 0.459 & 9 \\ \hline
07    &                           SL                       & 0.364 & 0.460 & 0.640 & 0.556 & 0.409 & 0.812 & 8 \\ \hline
08    &                           MC                       & 0.313 & 0.532 & 0.092 & 0.255 & 0.971 & 0.605 & 7 \\ \hline
09    &                           SC                       & 0.382 & 0.483 & 0.941 & 0.977 & 0.885 & 0.984 & 9 \\ \hline
10    &                           MLD                      & 0.343 & 2.907 & 0.283 & 0.991 & 0.903 & 0.060 & 5 \\ \hline
\end{tabular}
\begin{tablenotes}[para,flushleft]
M = mono, L = line, S = stereo, C = clothoid, MLD = \cite{hur2013multilane}
\end{tablenotes}
\end{threeparttable}
\end{table}

As further research is required for this problem, and to allow future researchers to compare their work with ours, we published our datasets and the associated GT values online\footnote{The dataset and the annotations are available on our lab's website: http://www.ira.disco.unimib.it/ego-lane-estimation-by-modeling-lanes-and-sensor-failures}.

\begin{figure*}
  \centering   
    \subfloat[]{\includegraphics[width=1.00\textwidth]{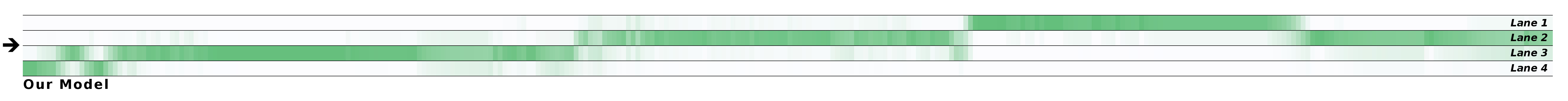}\label{fig:lanetransition01-a}}\hspace{\fill}\\
    \subfloat[]{\includegraphics[width=1.00\textwidth]{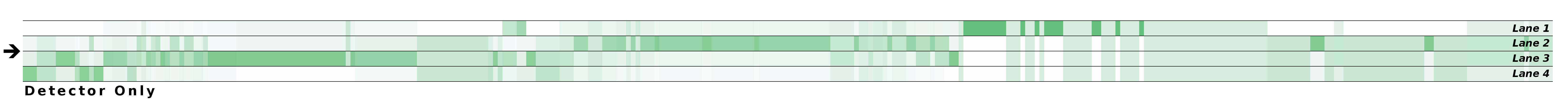}\label{fig:lanetransition01-b}}\hspace{\fill}\\
    \subfloat[]{\includegraphics[width=1.00\textwidth]{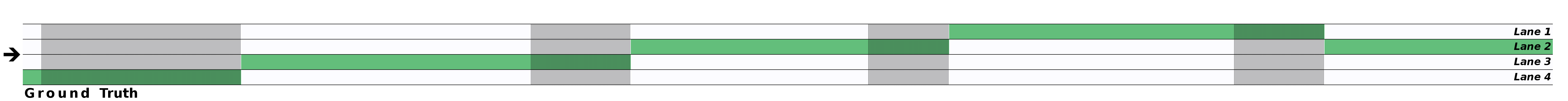}\label{fig:lanetransition01-c}}\hspace{\fill}\\
    \caption{A short section of	 the 4-lanes A4 highway in Italy. More saturated colors correspond to a higher probability of being in a lane. The figure presents a comparison between our model (a) \wrt the results achieved using the detector only (b). (c) is the GT, in grey are the transitions between lanes. Our proposal yields good improvements, with more stable detections, \wrt the detector's results.}
  \label{fig:lanetransition01}
\end{figure*}

\section{Results}\label{sec:discussion}
We evaluated the localization performances of our proposal comparing the ego-lane estimates \wrt the GT, for a set of configurations (with / without the proposed model, line detector, etc.). The results are presented in the Tables, on a per-frame basis, reporting whether correct lane classifications were achieved.

\begin{table}[]
\centering
\begin{threeparttable}
\caption{Run\#1 - Detector Only}
\label{tab:run01detector}
\begin{tabular}{lc|l|l|l|l|l}
\cline{3-6}
                                                       & \multicolumn{1}{l|}{}           & \multicolumn{4}{c|}{Actual Class}                                                                                     &                             \\ \cline{3-7} 
                                                       & \multicolumn{1}{l|}{}           & \multicolumn{1}{c|}{Lane 1} & \multicolumn{1}{c|}{Lane 2} & \multicolumn{1}{c|}{Lane 3} & \multicolumn{1}{c|}{Lane 4} & \multicolumn{1}{c|}{Total}  \\ \hline
\multicolumn{1}{|l|}{\multirow{4}{*}{\rotatebox[origin=c]{90}{\parbox[c]{1.6cm}{\centering Predicted Class}}}} & Lane 1                          & 1781                      & 21                      & 5                      & 1                      & \multicolumn{1}{l|}{1808} \\ \cline{2-7} 
\multicolumn{1}{|l|}{}                                 & Lane 2                          & 59                      & 1613                      & 190                      & 46                      & \multicolumn{1}{l|}{1908} \\ \cline{2-7} 
\multicolumn{1}{|l|}{}                                 & Lane 3                          & 147                      & 42                      & 730                      & 166                      & \multicolumn{1}{l|}{1085} \\ \cline{2-7} 
\multicolumn{1}{|l|}{}                                 & Lane 4                          & 1                      & 21                      & 17                      & 215                      & \multicolumn{1}{l|}{254} \\ \cline{2-7} 
\multicolumn{1}{|l|}{}                                 & \multicolumn{1}{l|}{Unassigned} & 196                      & 754                      & 1310                      & 456                      & \multicolumn{1}{l|}{2716} \\ \hline
\multicolumn{1}{l|}{}                                  & Support                         & 2184                      & 2451                      & 2252                      & 884                      & \multicolumn{1}{l|}{7771} \\ \cline{2-7} 
\end{tabular}
\begin{tablenotes}[para,flushleft]
Mean Precision: 0.83 Mean Recall: 0.51 Mean F1: 0.61 Accuracy 0.55 Logloss: 0.99
\end{tablenotes}
\end{threeparttable}
\end{table}
\begin{table}[]
\centering
\begin{threeparttable}
\caption{Run\#1 - Model}
\label{tab:run01model}
\begin{tabular}{lc|l|l|l|l|l}
\cline{3-6}
                                                       & \multicolumn{1}{l|}{}           & \multicolumn{4}{c|}{Actual Class}                                                                                     &                             \\ \cline{3-7} 
                                                       & \multicolumn{1}{l|}{}           & \multicolumn{1}{c|}{Lane 1} & \multicolumn{1}{c|}{Lane 2} & \multicolumn{1}{c|}{Lane 3} & \multicolumn{1}{c|}{Lane 4} & \multicolumn{1}{c|}{Total}  \\ \hline
\multicolumn{1}{|l|}{\multirow{4}{*}{\rotatebox[origin=c]{90}{\parbox[c]{1.6cm}{\centering Predicted Class}}}} & Lane 1                          & 2079                      & 27                      & 0                      & 0                      & \multicolumn{1}{l|}{2106} \\ \cline{2-7} 
\multicolumn{1}{|l|}{}                                 & Lane 2                          & 67                      & 2284                      & 346                      & 67                      & \multicolumn{1}{l|}{2764} \\ \cline{2-7} 
\multicolumn{1}{|l|}{}                                 & Lane 3                          & 17                      & 117                      & 1853                      & 409                      & \multicolumn{1}{l|}{2396} \\ \cline{2-7} 
\multicolumn{1}{|l|}{}                                 & Lane 4                          & 21                      & 23                      & 53                      & 405                      & \multicolumn{1}{l|}{502} \\ \cline{2-7} 
\multicolumn{1}{|l|}{}                                 & \multicolumn{1}{l|}{Unassigned} & 0                      & 0                      & 0                      & 3                      & \multicolumn{1}{l|}{3} \\ \hline
\multicolumn{1}{l|}{}                                  & Support                         & 2184                      & 2451                      & 2252                      & 884                      & \multicolumn{1}{l|}{7771} \\ \cline{2-7} 
\end{tabular}
\begin{tablenotes}[para,flushleft]
Mean Precision: 0.84 Mean Recall: 0.79 Mean F1: 0.80 Accuracy 0.85 Logloss: 0.51
\end{tablenotes}
\end{threeparttable}
\end{table}

\begin{table}[]
\centering
\begin{threeparttable}
\caption{Run\#2 - Detector Only}
\label{tab:run02detector}
\begin{tabular}{lc|l|l|l|l|l}
\cline{3-6}
                                                       & \multicolumn{1}{l|}{}           & \multicolumn{4}{c|}{Actual Class}                                                                                     &                             \\ \cline{3-7} 
                                                       & \multicolumn{1}{l|}{}           & \multicolumn{1}{c|}{Lane 1} & \multicolumn{1}{c|}{Lane 2} & \multicolumn{1}{c|}{Lane 3} & \multicolumn{1}{c|}{Lane 4} & \multicolumn{1}{c|}{Total}  \\ \hline
\multicolumn{1}{|l|}{\multirow{4}{*}{\rotatebox[origin=c]{90}{\parbox[c]{1.6cm}{\centering Predicted Class}}}} & Lane 1                          & 1941                      & 31                      & 2                      & 4                      & \multicolumn{1}{l|}{1978} \\ \cline{2-7} 
\multicolumn{1}{|l|}{}                                 & Lane 2                          & 9                      & 1611                      & 147                      & 31                      & \multicolumn{1}{l|}{1798} \\ \cline{2-7} 
\multicolumn{1}{|l|}{}                                 & Lane 3                          & 6                      & 21                      & 762                      & 171                      & \multicolumn{1}{l|}{960} \\ \cline{2-7} 
\multicolumn{1}{|l|}{}                                 & Lane 4                          & 3                      & 3                      & 5                      & 223                      & \multicolumn{1}{l|}{234} \\ \cline{2-7} 
\multicolumn{1}{|l|}{}                                 & \multicolumn{1}{l|}{Unassigned} & 225                      & 785                      & 1336                      & 455                      & \multicolumn{1}{l|}{2801} \\ \hline
\multicolumn{1}{l|}{}                                  & Support                         & 2184                      & 2451                      & 2252                      & 884                      & \multicolumn{1}{l|}{7771} \\ \cline{2-7} 
\end{tabular}
\begin{tablenotes}[para,flushleft]
Mean Precision: 0.90 Mean Recall: 0.53 Mean F1: 0.64 Accuracy 0.58 Logloss: 0.93
\end{tablenotes}
\end{threeparttable}
\end{table}
\begin{table}[]
\centering
\begin{threeparttable}
\caption{Run\#2 - Model}
\label{tab:run02model}
\begin{tabular}{lc|l|l|l|l|l}
\cline{3-6}
                                                       & \multicolumn{1}{l|}{}           & \multicolumn{4}{c|}{Actual Class}                                                                                     &                             \\ \cline{3-7} 
                                                       & \multicolumn{1}{l|}{}           & \multicolumn{1}{c|}{Lane 1} & \multicolumn{1}{c|}{Lane 2} & \multicolumn{1}{c|}{Lane 3} & \multicolumn{1}{c|}{Lane 4} & \multicolumn{1}{c|}{Total}  \\ \hline
\multicolumn{1}{|l|}{\multirow{4}{*}{\rotatebox[origin=c]{90}{\parbox[c]{1.6cm}{\centering Predicted Class}}}} & Lane 1                          & 2117                      & 94                      & 2                      & 0                      & \multicolumn{1}{l|}{2213} \\ \cline{2-7} 
\multicolumn{1}{|l|}{}                                 & Lane 2                          & 64                      & 2288                      & 352                      & 49                      & \multicolumn{1}{l|}{2753} \\ \cline{2-7} 
\multicolumn{1}{|l|}{}                                 & Lane 3                          & 3                      & 69                      & 1883                      & 385                      & \multicolumn{1}{l|}{2340} \\ \cline{2-7} 
\multicolumn{1}{|l|}{}                                 & Lane 4                          & 0                      & 0                      & 15                      & 450                      & \multicolumn{1}{l|}{465} \\ \cline{2-7} 
\multicolumn{1}{|l|}{}                                 & \multicolumn{1}{l|}{Unassigned} & 0                      & 0                      & 0                      & 0                      & \multicolumn{1}{l|}{0} \\ \hline
\multicolumn{1}{l|}{}                                  & Support                         & 2184                      & 2451                      & 2252                      & 884                      & \multicolumn{1}{l|}{7771} \\ \cline{2-7} 
\end{tabular}
\begin{tablenotes}[para,flushleft]
Mean Precision: 0.89 Mean Recall: 0.81 Mean F1: 0.83 Accuracy 0.86 Logloss: 0.94
\end{tablenotes}
\end{threeparttable}
\end{table}

\begin{table}[]
\centering
\begin{threeparttable}
\caption{Run\#3 - Detector Only}
\label{tab:run03detector}
\begin{tabular}{lc|l|l|l|l|l}
\cline{3-6}
                                                       & \multicolumn{1}{l|}{}           & \multicolumn{4}{c|}{Actual Class}                                                                                     &                             \\ \cline{3-7} 
                                                       & \multicolumn{1}{l|}{}           & \multicolumn{1}{c|}{Lane 1} & \multicolumn{1}{c|}{Lane 2} & \multicolumn{1}{c|}{Lane 3} & \multicolumn{1}{c|}{Lane 4} & \multicolumn{1}{c|}{Total}  \\ \hline
\multicolumn{1}{|l|}{\multirow{4}{*}{\rotatebox[origin=c]{90}{\parbox[c]{1.6cm}{\centering Predicted Class}}}} & Lane 1                          & 1438                      & 30                      & 51                      & 0                      & \multicolumn{1}{l|}{1519} \\ \cline{2-7} 
\multicolumn{1}{|l|}{}                                 & Lane 2                          & 79                      & 1795                      & 238                      & 124                      & \multicolumn{1}{l|}{2236} \\ \cline{2-7} 
\multicolumn{1}{|l|}{}                                 & Lane 3                          & 350                      & 56                      & 1207                      & 269                      & \multicolumn{1}{l|}{1882} \\ \cline{2-7} 
\multicolumn{1}{|l|}{}                                 & Lane 4                          & 5                      & 47                      & 33                      & 217                      & \multicolumn{1}{l|}{302} \\ \cline{2-7} 
\multicolumn{1}{|l|}{}                                 & \multicolumn{1}{l|}{Unassigned} & 312                      & 523                      & 723                      & 274                      & \multicolumn{1}{l|}{1832} \\ \hline
\multicolumn{1}{l|}{}                                  & Support                         & 2184                      & 2451                      & 2252                      & 884                      & \multicolumn{1}{l|}{7771} \\ \cline{2-7} 
\end{tabular}
\begin{tablenotes}[para,flushleft]
Mean Precision: 0.77 Mean Recall: 0.54 Mean F1: 0.62 Accuracy 0.59 Logloss: 1.07
\end{tablenotes}
\end{threeparttable}
\end{table}
\begin{table}[]
\centering
\begin{threeparttable}
\caption{Run\#3 - Model}
\label{tab:run03model}
\begin{tabular}{lc|l|l|l|l|l}
\cline{3-6}
                                                       & \multicolumn{1}{l|}{}           & \multicolumn{4}{c|}{Actual Class}                                                                                     &                             \\ \cline{3-7} 
                                                       & \multicolumn{1}{l|}{}           & \multicolumn{1}{c|}{Lane 1} & \multicolumn{1}{c|}{Lane 2} & \multicolumn{1}{c|}{Lane 3} & \multicolumn{1}{c|}{Lane 4} & \multicolumn{1}{c|}{Total}  \\ \hline
\multicolumn{1}{|l|}{\multirow{4}{*}{\rotatebox[origin=c]{90}{\parbox[c]{1.6cm}{\centering Predicted Class}}}} & Lane 1                          & 1967                      & 69                      & 48                      & 1                      & \multicolumn{1}{l|}{2085} \\ \cline{2-7} 
\multicolumn{1}{|l|}{}                                 & Lane 2                          & 85                      & 2268                      & 313                      & 70                      & \multicolumn{1}{l|}{2736} \\ \cline{2-7} 
\multicolumn{1}{|l|}{}                                 & Lane 3                          & 104                      & 99                      & 1817                      & 317                      & \multicolumn{1}{l|}{2337} \\ \cline{2-7} 
\multicolumn{1}{|l|}{}                                 & Lane 4                          & 28                      & 15                      & 74                      & 496                      & \multicolumn{1}{l|}{613} \\ \cline{2-7} 
\multicolumn{1}{|l|}{}                                 & \multicolumn{1}{l|}{Unassigned} & 0                      & 0                      & 0                      & 0                      & \multicolumn{1}{l|}{0} \\ \hline
\multicolumn{1}{l|}{}                                  & Support                         & 2184                      & 2451                      & 2252                      & 884                      & \multicolumn{1}{l|}{7771} \\ \cline{2-7} 
\end{tabular}
\begin{tablenotes}[para,flushleft]
Mean Precision: 0.83 Mean Recall: 0.79 Mean F1: 0.81 Accuracy 0.84 Logloss: 0.63
\end{tablenotes}
\end{threeparttable}
\end{table}

\begin{table}[]
\centering
\begin{threeparttable}
\caption{Run\#4 - Detector Only}
\label{tab:run04detector}
\begin{tabular}{lc|l|l|l|l|l}
\cline{3-6}
                                                       & \multicolumn{1}{l|}{}           & \multicolumn{4}{c|}{Actual Class}                                                                                     &                             \\ \cline{3-7} 
                                                       & \multicolumn{1}{l|}{}           & \multicolumn{1}{c|}{Lane 1} & \multicolumn{1}{c|}{Lane 2} & \multicolumn{1}{c|}{Lane 3} & \multicolumn{1}{c|}{Lane 4} & \multicolumn{1}{c|}{Total}  \\ \hline
\multicolumn{1}{|l|}{\multirow{4}{*}{\rotatebox[origin=c]{90}{\parbox[c]{1.6cm}{\centering Predicted Class}}}} & Lane 1                          & 1596                      & 27                      & 12                      & 8                      & \multicolumn{1}{l|}{1643} \\ \cline{2-7} 
\multicolumn{1}{|l|}{}                                 & Lane 2                          & 20                      & 1910                      & 190                      & 20                      & \multicolumn{1}{l|}{2140} \\ \cline{2-7} 
\multicolumn{1}{|l|}{}                                 & Lane 3                          & 11                      & 27                      & 1283                      & 327                      & \multicolumn{1}{l|}{1648} \\ \cline{2-7} 
\multicolumn{1}{|l|}{}                                 & Lane 4                          & 7                      & 0                      & 6                      & 206                      & \multicolumn{1}{l|}{219} \\ \cline{2-7} 
\multicolumn{1}{|l|}{}                                 & \multicolumn{1}{l|}{Unassigned} & 550                      & 487                      & 761                      & 323                      & \multicolumn{1}{l|}{2121} \\ \hline
\multicolumn{1}{l|}{}                                  & Support                         & 2184                      & 2451                      & 2252                      & 884                      & \multicolumn{1}{l|}{7771} \\ \cline{2-7} 
\end{tabular}
\begin{tablenotes}[para,flushleft]
Mean Precision: 0.89 Mean Recall: 0.57 Mean F1: 0.67 Accuracy 0.64 Logloss: 0.80
\end{tablenotes}
\end{threeparttable}
\end{table}
\begin{table}[]
\centering
\begin{threeparttable}
\caption{Run\#4 - Model}
\label{tab:run04model}
\begin{tabular}{lc|l|l|l|l|l}
\cline{3-6}
                                                       & \multicolumn{1}{l|}{}           & \multicolumn{4}{c|}{Actual Class}                                                                                     &                             \\ \cline{3-7} 
                                                       & \multicolumn{1}{l|}{}           & \multicolumn{1}{c|}{Lane 1} & \multicolumn{1}{c|}{Lane 2} & \multicolumn{1}{c|}{Lane 3} & \multicolumn{1}{c|}{Lane 4} & \multicolumn{1}{c|}{Total}  \\ \hline
\multicolumn{1}{|l|}{\multirow{4}{*}{\rotatebox[origin=c]{90}{\parbox[c]{1.6cm}{\centering Predicted Class}}}} & Lane 1                          & 2095                      & 38                      & 0                      & 0                      & \multicolumn{1}{l|}{2133} \\ \cline{2-7} 
\multicolumn{1}{|l|}{}                                 & Lane 2                          & 86                      & 2286                      & 240                      & 17                      & \multicolumn{1}{l|}{2629} \\ \cline{2-7} 
\multicolumn{1}{|l|}{}                                 & Lane 3                          & 3                      & 127                      & 1907                      & 306                      & \multicolumn{1}{l|}{2343} \\ \cline{2-7} 
\multicolumn{1}{|l|}{}                                 & Lane 4                          & 0                      & 0                      & 105                      & 552                      & \multicolumn{1}{l|}{657} \\ \cline{2-7} 
\multicolumn{1}{|l|}{}                                 & \multicolumn{1}{l|}{Unassigned} & 0                      & 0                      & 0                      & 9                      & \multicolumn{1}{l|}{9} \\ \hline
\multicolumn{1}{l|}{}                                  & Support                         & 2184                      & 2451                      & 2252                      & 884                      & \multicolumn{1}{l|}{7771} \\ \cline{2-7} 
\end{tabular}
\begin{tablenotes}[para,flushleft]
Mean Precision: 0.87 Mean Recall: 0.84 Mean F1: 0.85 Accuracy 0.88 Logloss: 0.49
\end{tablenotes}
\end{threeparttable}
\end{table}

\begin{table}[]
\centering
\begin{threeparttable}
\caption{Run\#5 - Detector Only}
\label{tab:run05detector}
\begin{tabular}{lc|l|l|l|l|l}
\cline{3-6}
                                                       & \multicolumn{1}{l|}{}           & \multicolumn{4}{c|}{Actual Class}                                                                                     &                             \\ \cline{3-7} 
                                                       & \multicolumn{1}{l|}{}           & \multicolumn{1}{c|}{Lane 1} & \multicolumn{1}{c|}{Lane 2} & \multicolumn{1}{c|}{Lane 3} & \multicolumn{1}{c|}{Lane 4} & \multicolumn{1}{c|}{Total}  \\ \hline
\multicolumn{1}{|l|}{\multirow{4}{*}{\rotatebox[origin=c]{90}{\parbox[c]{1.6cm}{\centering Predicted Class}}}} & Lane 1                          & 0                      & 0                      & 0                      & 0                      & \multicolumn{1}{l|}{0} \\ \cline{2-7} 
\multicolumn{1}{|l|}{}                                 & Lane 2                          & 1                      & 0                      & 45                      & 0                      & \multicolumn{1}{l|}{46} \\ \cline{2-7} 
\multicolumn{1}{|l|}{}                                 & Lane 3                          & 39                      & 0                      & 6                      & 0                      & \multicolumn{1}{l|}{45} \\ \cline{2-7} 
\multicolumn{1}{|l|}{}                                 & Lane 4                          & 0                      & 0                      & 0                      & 0                      & \multicolumn{1}{l|}{0} \\ \cline{2-7} 
\multicolumn{1}{|l|}{}                                 & \multicolumn{1}{l|}{Unassigned} & 2122                      & 2413                      & 2171                      & 876                      & \multicolumn{1}{l|}{7582} \\ \hline
\multicolumn{1}{l|}{}                                  & Support                         & 2162                      & 2413                      & 2222                      & 876                      & \multicolumn{1}{l|}{7673} \\ \cline{2-7} 
\end{tabular}
\begin{tablenotes}[para,flushleft]
Mean Precision: 0.03 Mean Recall: 0.00 Mean F1: 0.00 Accuracy 0.00 Logloss: 1.80
\end{tablenotes}
\end{threeparttable}
\end{table}
\begin{table}[]
\centering
\begin{threeparttable}
\caption{Run\#5 - Model}
\label{tab:run05model}
\begin{tabular}{lc|l|l|l|l|l}
\cline{3-6}
                                                       & \multicolumn{1}{l|}{}           & \multicolumn{4}{c|}{Actual Class}                                                                                     &                             \\ \cline{3-7} 
                                                       & \multicolumn{1}{l|}{}           & \multicolumn{1}{c|}{Lane 1} & \multicolumn{1}{c|}{Lane 2} & \multicolumn{1}{c|}{Lane 3} & \multicolumn{1}{c|}{Lane 4} & \multicolumn{1}{c|}{Total}  \\ \hline
\multicolumn{1}{|l|}{\multirow{4}{*}{\rotatebox[origin=c]{90}{\parbox[c]{1.6cm}{\centering Predicted Class}}}} & Lane 1                          & 1120                      & 549                      & 163                      & 59                      & \multicolumn{1}{l|}{1891} \\ \cline{2-7} 
\multicolumn{1}{|l|}{}                                 & Lane 2                          & 448                      & 1320                      & 812                      & 160                      & \multicolumn{1}{l|}{2740} \\ \cline{2-7} 
\multicolumn{1}{|l|}{}                                 & Lane 3                          & 576                      & 407                      & 1037                      & 515                      & \multicolumn{1}{l|}{2535} \\ \cline{2-7} 
\multicolumn{1}{|l|}{}                                 & Lane 4                          & 18                      & 137                      & 210                      & 142                      & \multicolumn{1}{l|}{507} \\ \cline{2-7} 
\multicolumn{1}{|l|}{}                                 & \multicolumn{1}{l|}{Unassigned} & 0                      & 0                      & 0                      & 0                      & \multicolumn{1}{l|}{0} \\ \hline
\multicolumn{1}{l|}{}                                  & Support                         & 2162                      & 2413                      & 2222                      & 876                      & \multicolumn{1}{l|}{7673} \\ \cline{2-7} 
\end{tabular}
\begin{tablenotes}[para,flushleft]
Mean Precision: 0.44 Mean Recall: 0.42 Mean F1: 0.42 Accuracy 0.47 Logloss: 1.26
\end{tablenotes}
\end{threeparttable}
\end{table}

\begin{table}[]
\centering
\begin{threeparttable}
\caption{Run\#6 - Detector Only}
\label{tab:run06detector}
\begin{tabular}{lc|l|l|l|l}
\cline{3-5}
                                                       & \multicolumn{1}{l|}{}          & \multicolumn{3}{c|}{Actual Class}                                                       &                             \\ \cline{3-6} 
                                                       & \multicolumn{1}{l|}{\textit{}} & \multicolumn{1}{c|}{Lane 1} & \multicolumn{1}{c|}{Lane 2} & \multicolumn{1}{c|}{Lane 3} & \multicolumn{1}{c|}{Total}  \\ \hline
\multicolumn{1}{|l|}{\multirow{4}{*}{\rotatebox[origin=c]{90}{\parbox[c]{1.2cm}{\centering Predicted Class}}}} & Lane 1                         & 1405                      & 162                      & 2                      & \multicolumn{1}{l|}{1569} \\ \cline{2-6} 
\multicolumn{1}{|l|}{}                                 & Lane 2                         & 68                      & 3004                      & 275                      & \multicolumn{1}{l|}{3347} \\ \cline{2-6} 
\multicolumn{1}{|l|}{}                                 & Lane 3                         & 27                      & 6                      & 886                      & \multicolumn{1}{l|}{919} \\ \cline{2-6} 
\multicolumn{1}{|l|}{}                                 & Unassigned                     & 479                      & 424                      & 1132                      & \multicolumn{1}{l|}{2035} \\ \hline
\multicolumn{1}{l|}{}                                  & Support                        & 1979                      & 3596                      & 2295                      & \multicolumn{1}{l|}{7870} \\ \cline{2-6} 
\end{tabular}
\begin{tablenotes}[para,flushleft]
Mean Precision: 0.91 Mean Recall: 0.64 Mean F1: 0.73 Accuracy 0.67 Logloss: 0.73
\end{tablenotes}
\end{threeparttable}
\end{table}
\begin{table}[]
\centering
\begin{threeparttable}
\caption{Run\#6 - Model}
\label{tab:run06model}
\begin{tabular}{lc|l|l|l|l}
\cline{3-5}
                                                       & \multicolumn{1}{l|}{}          & \multicolumn{3}{c|}{Actual Class}                                                       &                             \\ \cline{3-6} 
                                                       & \multicolumn{1}{l|}{\textit{}} & \multicolumn{1}{c|}{Lane 1} & \multicolumn{1}{c|}{Lane 2} & \multicolumn{1}{c|}{Lane 3} & \multicolumn{1}{c|}{Total}  \\ \hline
\multicolumn{1}{|l|}{\multirow{4}{*}{\rotatebox[origin=c]{90}{\parbox[c]{1.2cm}{\centering Predicted Class}}}} & Lane 1                         & 1762                      & 263                      & 0                      & \multicolumn{1}{l|}{2025} \\ \cline{2-6} 
\multicolumn{1}{|l|}{}                                 & Lane 2                         & 150                      & 3323                      & 326                      & \multicolumn{1}{l|}{3799} \\ \cline{2-6} 
\multicolumn{1}{|l|}{}                                 & Lane 3                         & 66                      & 10                      & 1969                      & \multicolumn{1}{l|}{2045} \\ \cline{2-6} 
\multicolumn{1}{|l|}{}                                 & Unassigned                     & 1                      & 0                      & 0                      & \multicolumn{1}{l|}{1} \\ \hline
\multicolumn{1}{l|}{}                                  & Support                        & 1979                      & 3596                      & 2295                      & \multicolumn{1}{l|}{7870} \\ \cline{2-6} 
\end{tabular}
\begin{tablenotes}[para,flushleft]
Mean Precision: 0.90 Mean Recall: 0.89 Mean F1: 0.89 Accuracy 0.89 Logloss: 0.49
\end{tablenotes}
\end{threeparttable}
\end{table}

\begin{table}[]
\centering
\begin{threeparttable}
\caption{Run\#7 - Detector Only}
\label{tab:run07detector}
\begin{tabular}{lc|l|l|l|l}
\cline{3-5}
                                                       & \multicolumn{1}{l|}{}          & \multicolumn{3}{c|}{Actual Class}                                                       &                             \\ \cline{3-6} 
                                                       & \multicolumn{1}{l|}{\textit{}} & \multicolumn{1}{c|}{Lane 1} & \multicolumn{1}{c|}{Lane 2} & \multicolumn{1}{c|}{Lane 3} & \multicolumn{1}{c|}{Total}  \\ \hline
\multicolumn{1}{|l|}{\multirow{4}{*}{\rotatebox[origin=c]{90}{\parbox[c]{1.2cm}{\centering Predicted Class}}}} & Lane 1                         & 1380                      & 163                      & 2                      & \multicolumn{1}{l|}{1545} \\ \cline{2-6} 
\multicolumn{1}{|l|}{}                                 & Lane 2                         & 69                      & 2850                      & 216                      & \multicolumn{1}{l|}{3135} \\ \cline{2-6} 
\multicolumn{1}{|l|}{}                                 & Lane 3                         & 29                      & 19                      & 901                      & \multicolumn{1}{l|}{949} \\ \cline{2-6} 
\multicolumn{1}{|l|}{}                                 & Unassigned                     & 501                      & 564                      & 1176                      & \multicolumn{1}{l|}{2241} \\ \hline
\multicolumn{1}{l|}{}                                  & Support                        & 1979                      & 3596                      & 2295                      & \multicolumn{1}{l|}{7870} \\ \cline{2-6} 
\end{tabular}
\begin{tablenotes}[para,flushleft]
Mean Precision: 0.91 Mean Recall: 0.62 Mean F1: 0.72 Accuracy 0.65 Logloss: 0.73
\end{tablenotes}
\end{threeparttable}
\end{table}
\begin{table}[]
\centering
\begin{threeparttable}
\caption{Run\#7 - Model}
\label{tab:run07model}
\begin{tabular}{lc|l|l|l|l}
\cline{3-5}
                                                       & \multicolumn{1}{l|}{}          & \multicolumn{3}{c|}{Actual Class}                                                       &                             \\ \cline{3-6} 
                                                       & \multicolumn{1}{l|}{\textit{}} & \multicolumn{1}{c|}{Lane 1} & \multicolumn{1}{c|}{Lane 2} & \multicolumn{1}{c|}{Lane 3} & \multicolumn{1}{c|}{Total}  \\ \hline
\multicolumn{1}{|l|}{\multirow{4}{*}{\rotatebox[origin=c]{90}{\parbox[c]{1.2cm}{\centering Predicted Class}}}} & Lane 1                         & 1749                      & 219                      & 0                      & \multicolumn{1}{l|}{1968} \\ \cline{2-6} 
\multicolumn{1}{|l|}{}                                 & Lane 2                         & 155                      & 3322                      & 278                      & \multicolumn{1}{l|}{3755} \\ \cline{2-6} 
\multicolumn{1}{|l|}{}                                 & Lane 3                         & 74                      & 55                      & 2017                      & \multicolumn{1}{l|}{2146} \\ \cline{2-6} 
\multicolumn{1}{|l|}{}                                 & Unassigned                     & 1                      & 0                      & 0                      & \multicolumn{1}{l|}{1} \\ \hline
\multicolumn{1}{l|}{}                                  & Support                        & 1979                      & 3596                      & 2295                      & \multicolumn{1}{l|}{7870} \\ \cline{2-6} 
\end{tabular}
\begin{tablenotes}[para,flushleft]
Mean Precision: 0.90 Mean Recall: 0.89 Mean F1: 0.89 Accuracy 0.90 Logloss: 0.48
\end{tablenotes}
\end{threeparttable}
\end{table}

\begin{table}[]
\centering
\begin{threeparttable}
\caption{Run\#8 - Detector Only}
\label{tab:run08detector}
\begin{tabular}{lc|l|l|l|l}
\cline{3-5}
                                                       & \multicolumn{1}{l|}{}          & \multicolumn{3}{c|}{Actual Class}                                                       &                             \\ \cline{3-6} 
                                                       & \multicolumn{1}{l|}{\textit{}} & \multicolumn{1}{c|}{Lane 1} & \multicolumn{1}{c|}{Lane 2} & \multicolumn{1}{c|}{Lane 3} & \multicolumn{1}{c|}{Total}  \\ \hline
\multicolumn{1}{|l|}{\multirow{4}{*}{\rotatebox[origin=c]{90}{\parbox[c]{1.2cm}{\centering Predicted Class}}}} & Lane 1                         & 1263                      & 82                      & 3                      & \multicolumn{1}{l|}{1348} \\ \cline{2-6} 
\multicolumn{1}{|l|}{}                                 & Lane 2                         & 57                      & 2527                      & 208                      & \multicolumn{1}{l|}{2792} \\ \cline{2-6} 
\multicolumn{1}{|l|}{}                                 & Lane 3                         & 69                      & 38                      & 748                      & \multicolumn{1}{l|}{855} \\ \cline{2-6} 
\multicolumn{1}{|l|}{}                                 & Unassigned                     & 590                      & 949                      & 1336                      & \multicolumn{1}{l|}{2875} \\ \hline
\multicolumn{1}{l|}{}                                  & Support                        & 1979                      & 3596                      & 2295                      & \multicolumn{1}{l|}{7870} \\ \cline{2-6} 
\end{tabular}
\begin{tablenotes}[para,flushleft]
Mean Precision: 0.90 Mean Recall: 0.55 Mean F1: 0.67 Accuracy 0.57 Logloss: 0.95
\end{tablenotes}
\end{threeparttable}
\end{table}
\begin{table}[]
\centering
\begin{threeparttable}
\caption{Run\#8 - Model}
\label{tab:run08model}
\begin{tabular}{lc|l|l|l|l}
\cline{3-5}
                                                       & \multicolumn{1}{l|}{}          & \multicolumn{3}{c|}{Actual Class}                                                       &                             \\ \cline{3-6} 
                                                       & \multicolumn{1}{l|}{\textit{}} & \multicolumn{1}{c|}{Lane 1} & \multicolumn{1}{c|}{Lane 2} & \multicolumn{1}{c|}{Lane 3} & \multicolumn{1}{c|}{Total}  \\ \hline
\multicolumn{1}{|l|}{\multirow{4}{*}{\rotatebox[origin=c]{90}{\parbox[c]{1.2cm}{\centering Predicted Class}}}} & Lane 1                         & 1671                      & 199                      & 49                      & \multicolumn{1}{l|}{1919} \\ \cline{2-6} 
\multicolumn{1}{|l|}{}                                 & Lane 2                         & 149                      & 3262                      & 565                      & \multicolumn{1}{l|}{3976} \\ \cline{2-6} 
\multicolumn{1}{|l|}{}                                 & Lane 3                         & 158                      & 135                      & 1681                      & \multicolumn{1}{l|}{1974} \\ \cline{2-6} 
\multicolumn{1}{|l|}{}                                 & Unassigned                     & 1                      & 0                      & 0                      & \multicolumn{1}{l|}{1} \\ \hline
\multicolumn{1}{l|}{}                                  & Support                        & 1979                      & 3596                      & 2295                      & \multicolumn{1}{l|}{7870} \\ \cline{2-6} 
\end{tabular}
\begin{tablenotes}[para,flushleft]
Mean Precision: 0.84 Mean Recall: 0.82 Mean F1: 0.83 Accuracy 0.84 Logloss: 0.64
\end{tablenotes}
\end{threeparttable}
\end{table}

\begin{table}[]
\centering
\begin{threeparttable}
\caption{Run\#9 - Detector Only}
\label{tab:run09detector}
\begin{tabular}{lc|l|l|l|l}
\cline{3-5}
                                                       & \multicolumn{1}{l|}{}          & \multicolumn{3}{c|}{Actual Class}                                                       &                             \\ \cline{3-6} 
                                                       & \multicolumn{1}{l|}{\textit{}} & \multicolumn{1}{c|}{Lane 1} & \multicolumn{1}{c|}{Lane 2} & \multicolumn{1}{c|}{Lane 3} & \multicolumn{1}{c|}{Total}  \\ \hline
\multicolumn{1}{|l|}{\multirow{4}{*}{\rotatebox[origin=c]{90}{\parbox[c]{1.2cm}{\centering Predicted Class}}}} & Lane 1                         & 1375                      & 90                      & 11                      & \multicolumn{1}{l|}{1476} \\ \cline{2-6} 
\multicolumn{1}{|l|}{}                                 & Lane 2                         & 46                      & 2882                      & 251                      & \multicolumn{1}{l|}{3179} \\ \cline{2-6} 
\multicolumn{1}{|l|}{}                                 & Lane 3                         & 55                      & 19                      & 824                      & \multicolumn{1}{l|}{898} \\ \cline{2-6} 
\multicolumn{1}{|l|}{}                                 & Unassigned                     & 503                      & 605                      & 1209                      & \multicolumn{1}{l|}{2317} \\ \hline
\multicolumn{1}{l|}{}                                  & Support                        & 1979                      & 3596                      & 2295                      & \multicolumn{1}{l|}{7870} \\ \cline{2-6} 
\end{tabular}
\begin{tablenotes}[para,flushleft]
Mean Precision: 0.91 Mean Recall: 0.61 Mean F1: 0.72 Accuracy 0.64 Logloss: 0.79
\end{tablenotes}
\end{threeparttable}
\end{table}
\begin{table}[]
\centering
\begin{threeparttable}
\caption{Run\#9 - Model}
\label{tab:run09model}
\begin{tabular}{lc|l|l|l|l}
\cline{3-5}
                                                       & \multicolumn{1}{l|}{}          & \multicolumn{3}{c|}{Actual Class}                                                       &                             \\ \cline{3-6} 
                                                       & \multicolumn{1}{l|}{\textit{}} & \multicolumn{1}{c|}{Lane 1} & \multicolumn{1}{c|}{Lane 2} & \multicolumn{1}{c|}{Lane 3} & \multicolumn{1}{c|}{Total}  \\ \hline
\multicolumn{1}{|l|}{\multirow{4}{*}{\rotatebox[origin=c]{90}{\parbox[c]{1.2cm}{\centering Predicted Class}}}} & Lane 1                         & 1777                      & 166                      & 10                      & \multicolumn{1}{l|}{1953} \\ \cline{2-6} 
\multicolumn{1}{|l|}{}                                 & Lane 2                         & 88                      & 3348                      & 363                      & \multicolumn{1}{l|}{3799} \\ \cline{2-6} 
\multicolumn{1}{|l|}{}                                 & Lane 3                         & 113                      & 82                      & 1922                      & \multicolumn{1}{l|}{2117} \\ \cline{2-6} 
\multicolumn{1}{|l|}{}                                 & Unassigned                     & 1                      & 0                      & 0                      & \multicolumn{1}{l|}{1} \\ \hline
\multicolumn{1}{l|}{}                                  & Support                        & 1979                      & 3596                      & 2295                      & \multicolumn{1}{l|}{7870} \\ \cline{2-6} 
\end{tabular}
\begin{tablenotes}[para,flushleft]
Mean Precision: 0.89 Mean Recall: 0.88 Mean F1: 0.89 Accuracy 0.89 Logloss: 0.56
\end{tablenotes}
\end{threeparttable}
\end{table}

\begin{table}[]
\centering
\begin{threeparttable}
\caption{Run\#10 - Detector Only}
\label{tab:run10detector}
\begin{tabular}{lc|l|l|l|l}
\cline{3-5}
                                                       & \multicolumn{1}{l|}{}          & \multicolumn{3}{c|}{Actual Class}                                                       &                             \\ \cline{3-6} 
                                                       & \multicolumn{1}{l|}{\textit{}} & \multicolumn{1}{c|}{Lane 1} & \multicolumn{1}{c|}{Lane 2} & \multicolumn{1}{c|}{Lane 3} & \multicolumn{1}{c|}{Total}  \\ \hline
\multicolumn{1}{|l|}{\multirow{4}{*}{\rotatebox[origin=c]{90}{\parbox[c]{1.2cm}{\centering Predicted Class}}}} & Lane 1                         & 153                      & 104                      & 0                      & \multicolumn{1}{l|}{257} \\ \cline{2-6} 
\multicolumn{1}{|l|}{}                                 & Lane 2                         & 58                      & 1714                      & 365                      & \multicolumn{1}{l|}{2137} \\ \cline{2-6} 
\multicolumn{1}{|l|}{}                                 & Lane 3                         & 72                      & 138                      & 203                      & \multicolumn{1}{l|}{413} \\ \cline{2-6} 
\multicolumn{1}{|l|}{}                                 & Unassigned                     & 1653                      & 1590                      & 1727                      & \multicolumn{1}{l|}{4970} \\ \hline
\multicolumn{1}{l|}{}                                  & Support                        & 1936                      & 3546                      & 2295                      & \multicolumn{1}{l|}{7777} \\ \cline{2-6} 
\end{tabular}
\begin{tablenotes}[para,flushleft]
Mean Precision: 0.62 Mean Recall: 0.21 Mean F1: 0.29 Accuracy 0.26 Logloss: 1.16
\end{tablenotes}
\end{threeparttable}
\end{table}
\begin{table}[]
\centering
\begin{threeparttable}
\caption{Run\#10 - Model}
\label{tab:run10model}
\begin{tabular}{lc|l|l|l|l}
\cline{3-5}
                                                       & \multicolumn{1}{l|}{}          & \multicolumn{3}{c|}{Actual Class}                                                       &                             \\ \cline{3-6} 
                                                       & \multicolumn{1}{l|}{\textit{}} & \multicolumn{1}{c|}{Lane 1} & \multicolumn{1}{c|}{Lane 2} & \multicolumn{1}{c|}{Lane 3} & \multicolumn{1}{c|}{Total}  \\ \hline
\multicolumn{1}{|l|}{\multirow{4}{*}{\rotatebox[origin=c]{90}{\parbox[c]{1.2cm}{\centering Predicted Class}}}} & Lane 1                         & 1436                      & 292                      & 0                      & \multicolumn{1}{l|}{1728} \\ \cline{2-6} 
\multicolumn{1}{|l|}{}                                 & Lane 2                         & 317                      & 2841                      & 326                      & \multicolumn{1}{l|}{3484} \\ \cline{2-6} 
\multicolumn{1}{|l|}{}                                 & Lane 3                         & 183                      & 411                      & 1969                      & \multicolumn{1}{l|}{2563} \\ \cline{2-6} 
\multicolumn{1}{|l|}{}                                 & Unassigned                     & 0                      & 2                      & 0                      & \multicolumn{1}{l|}{2} \\ \hline
\multicolumn{1}{l|}{}                                  & Support                        & 1936                      & 3546                      & 2295                      & \multicolumn{1}{l|}{7777} \\ \cline{2-6} 
\end{tabular}
\begin{tablenotes}[para,flushleft]
Mean Precision: 0.80 Mean Recall: 0.80 Mean F1: 0.80 Accuracy 0.80 Logloss: 1.08
\end{tablenotes}
\end{threeparttable}
\end{table}

\begin{figure}[]
	\centering        
	
	\subfloat[Run\#1 - Detector Only] {\includegraphics[width=0.48\columnwidth]{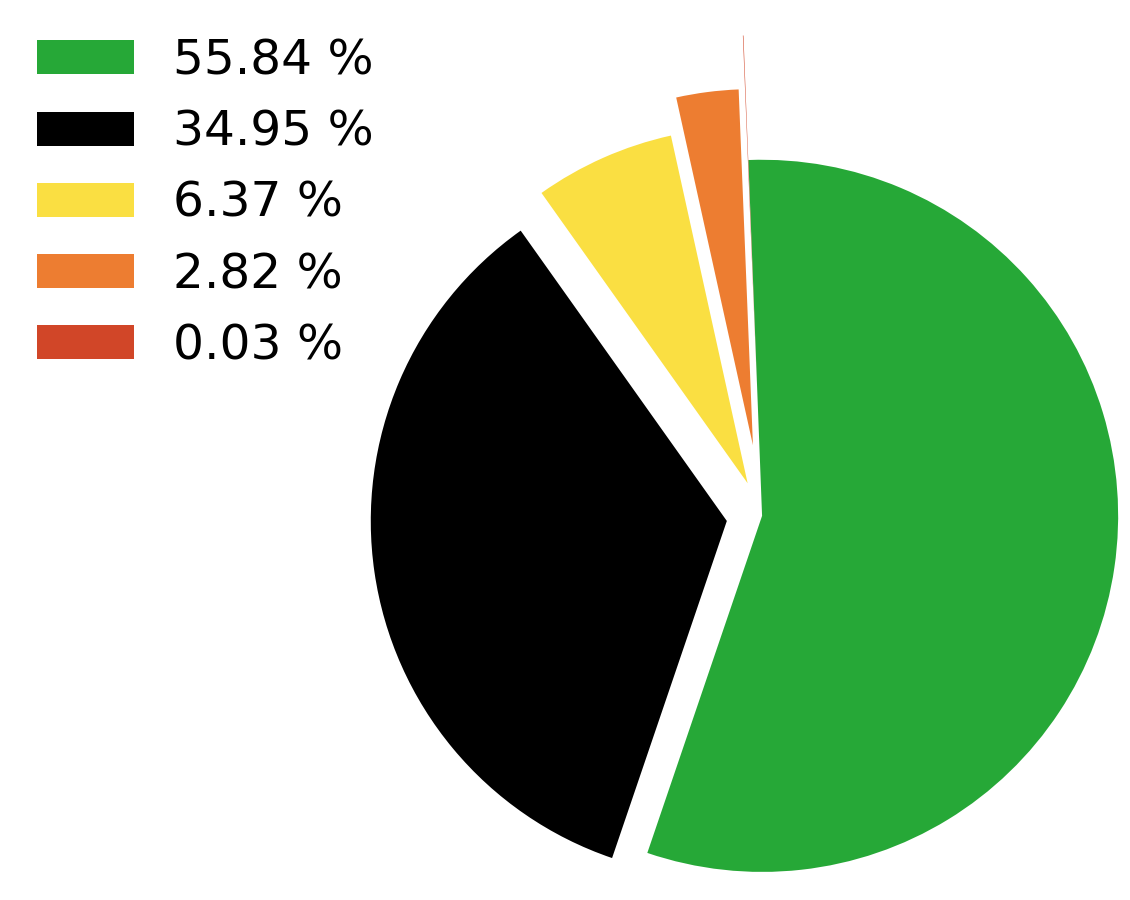}			\label{fig:graph1}}\hspace{\fill}
	\subfloat[Run\#1 - Model] {\includegraphics[width=0.48\columnwidth]{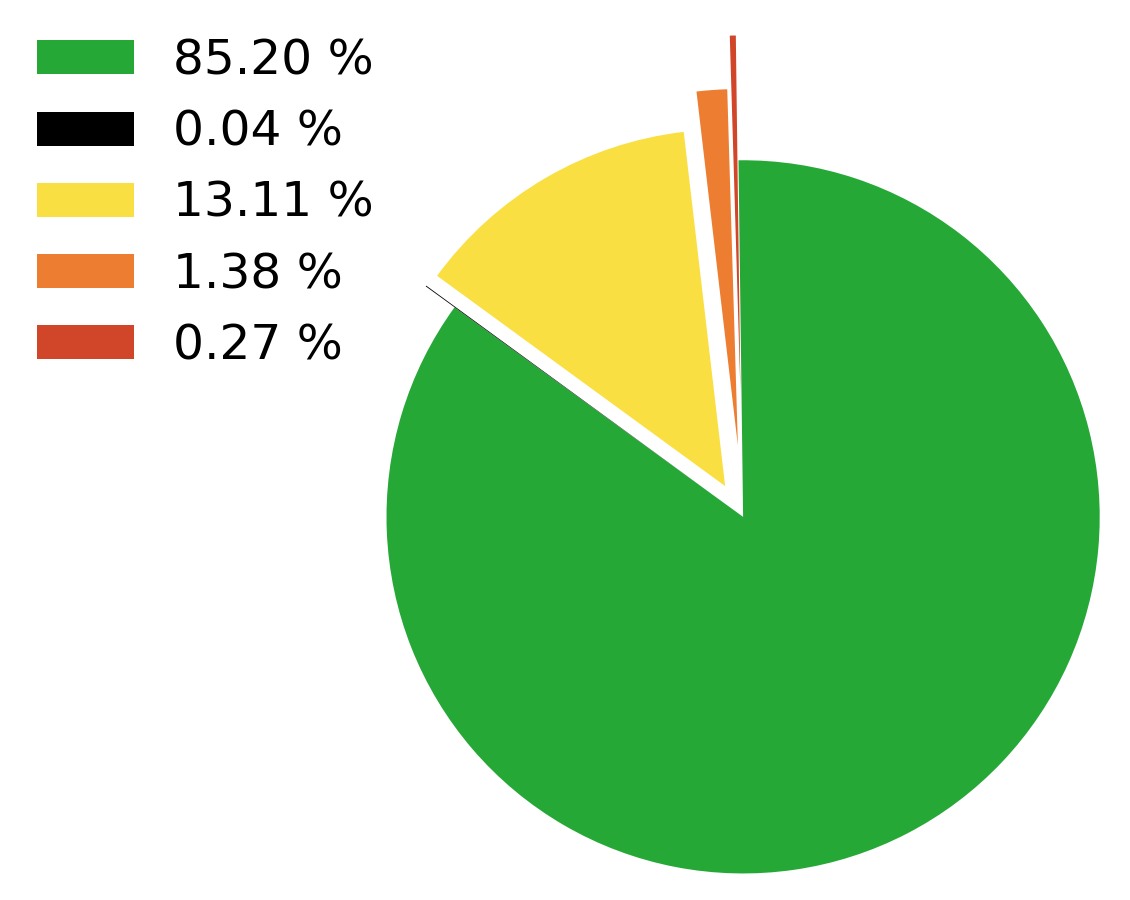}					\label{fig:graph2}}\\
	
	\subfloat[Run\#2 - Detector Only] {\includegraphics[width=0.48\columnwidth]{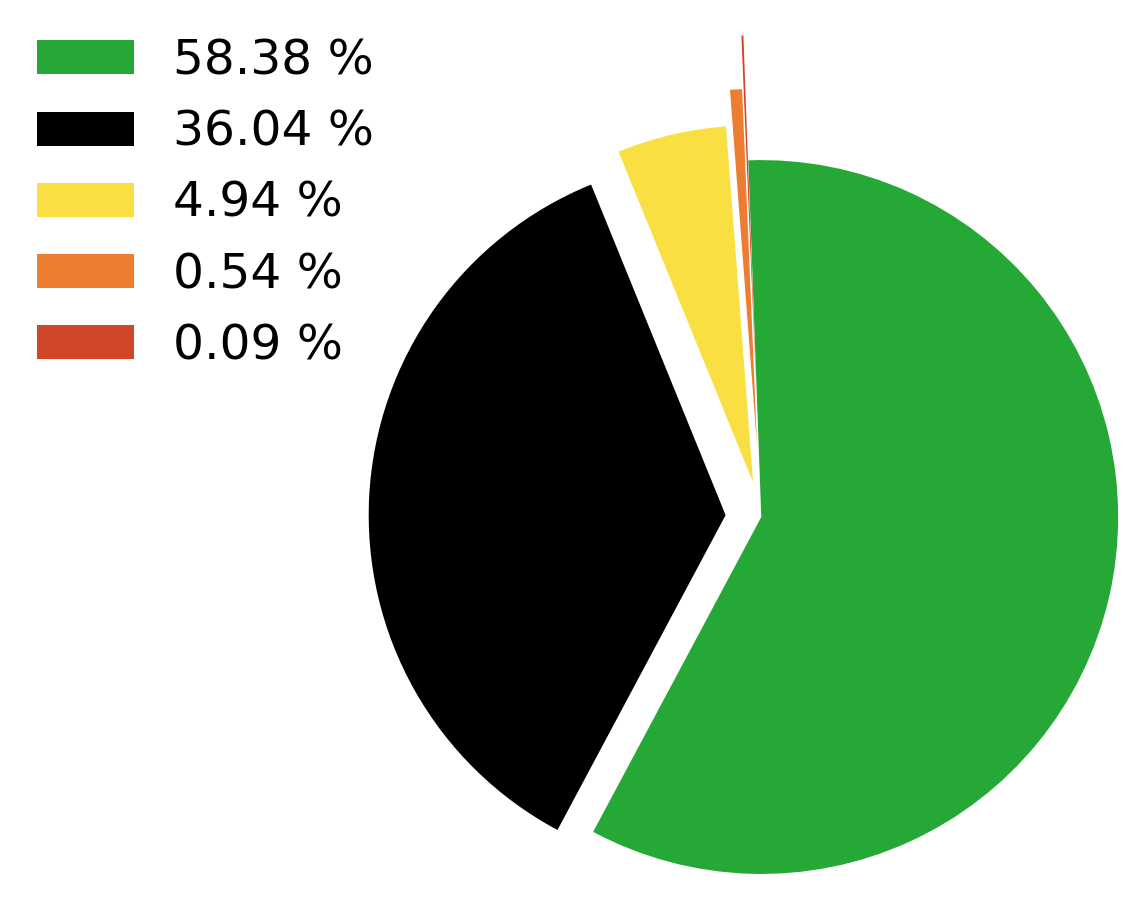}		\label{fig:graph3}}\hspace{\fill}
	\subfloat[Run\#2 - Model] {\includegraphics[width=0.48\columnwidth]{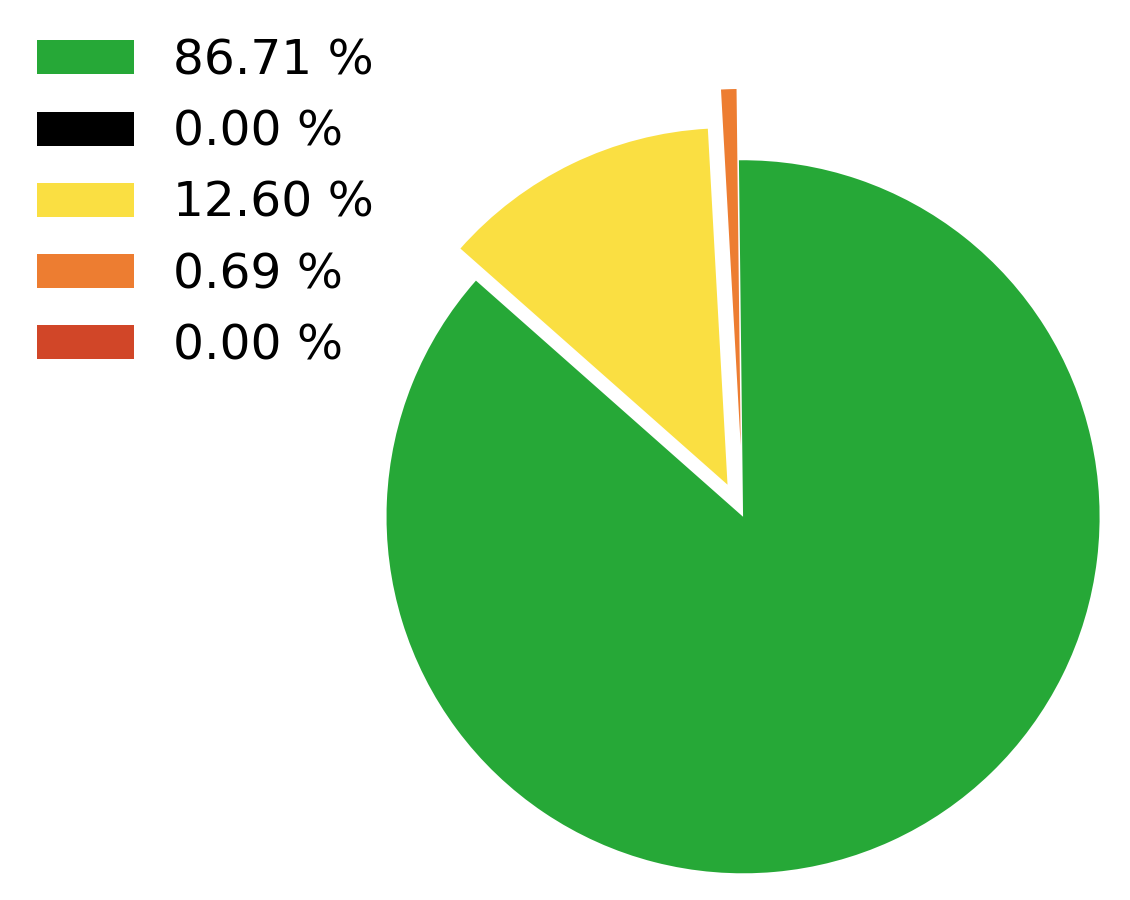}				\label{fig:graph4}}\\  
	
	\subfloat[Run\#3 - Detector Only]{\includegraphics[width=0.48\columnwidth]{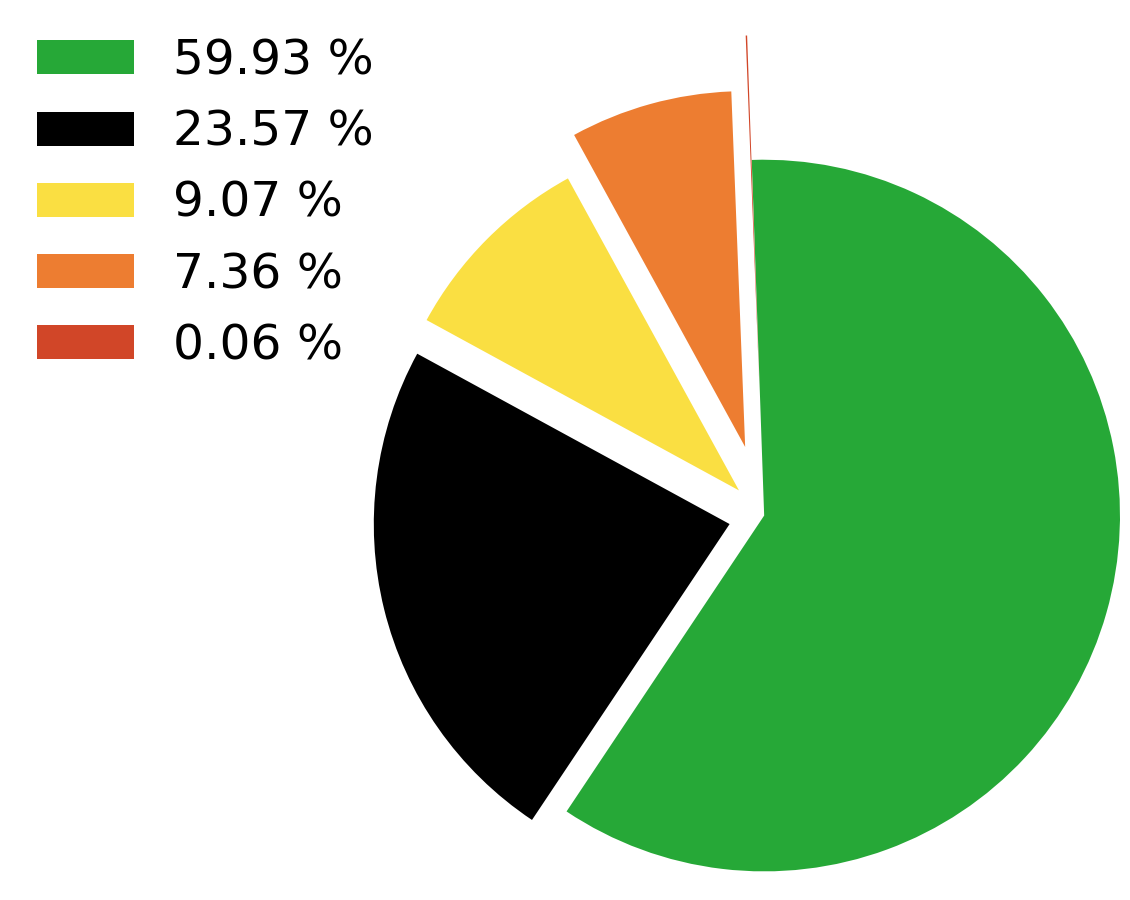}							\label{fig:graph15}}\hspace{\fill}
	\subfloat[Run\#3 - Model]{\includegraphics[width=0.48\columnwidth]{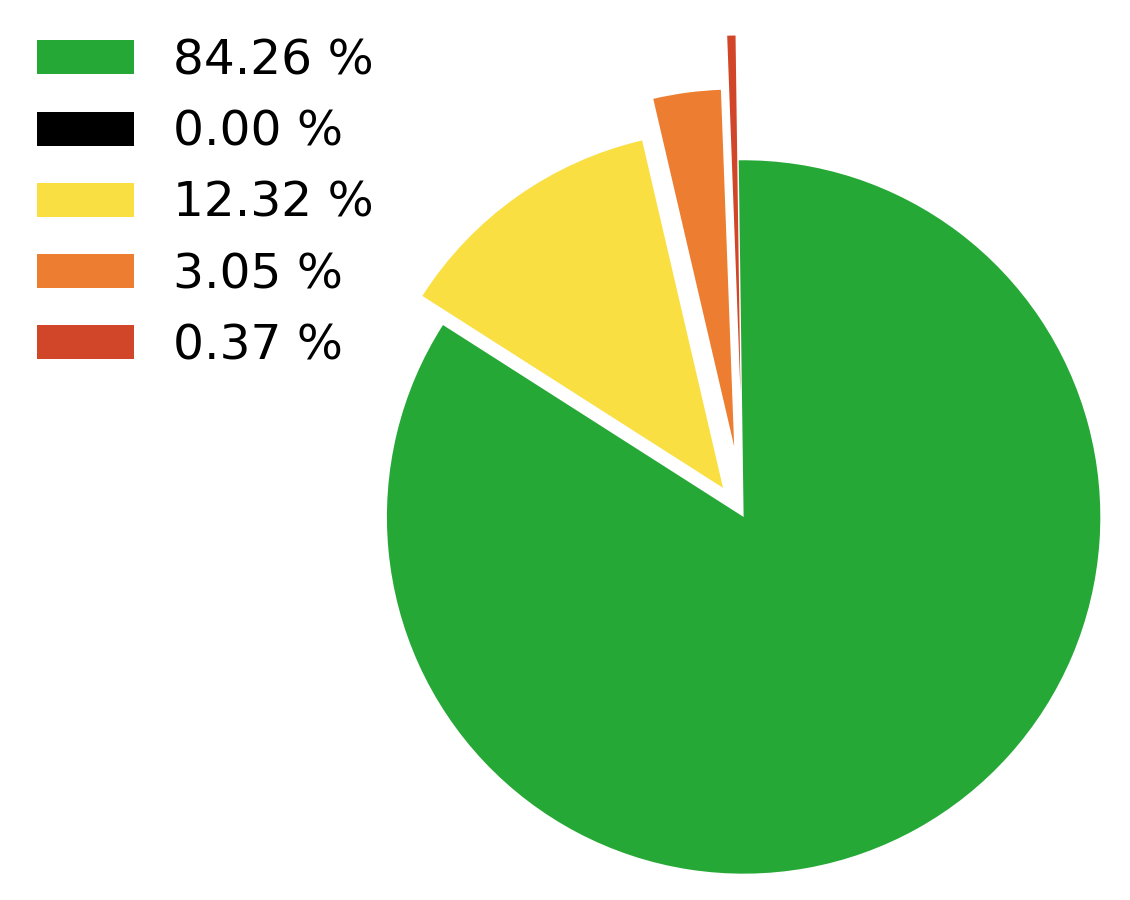}									
	\label{fig:graph16}}\\  
	
	\subfloat[Run\#4 - Detector Only]{\includegraphics[width=0.48\columnwidth]{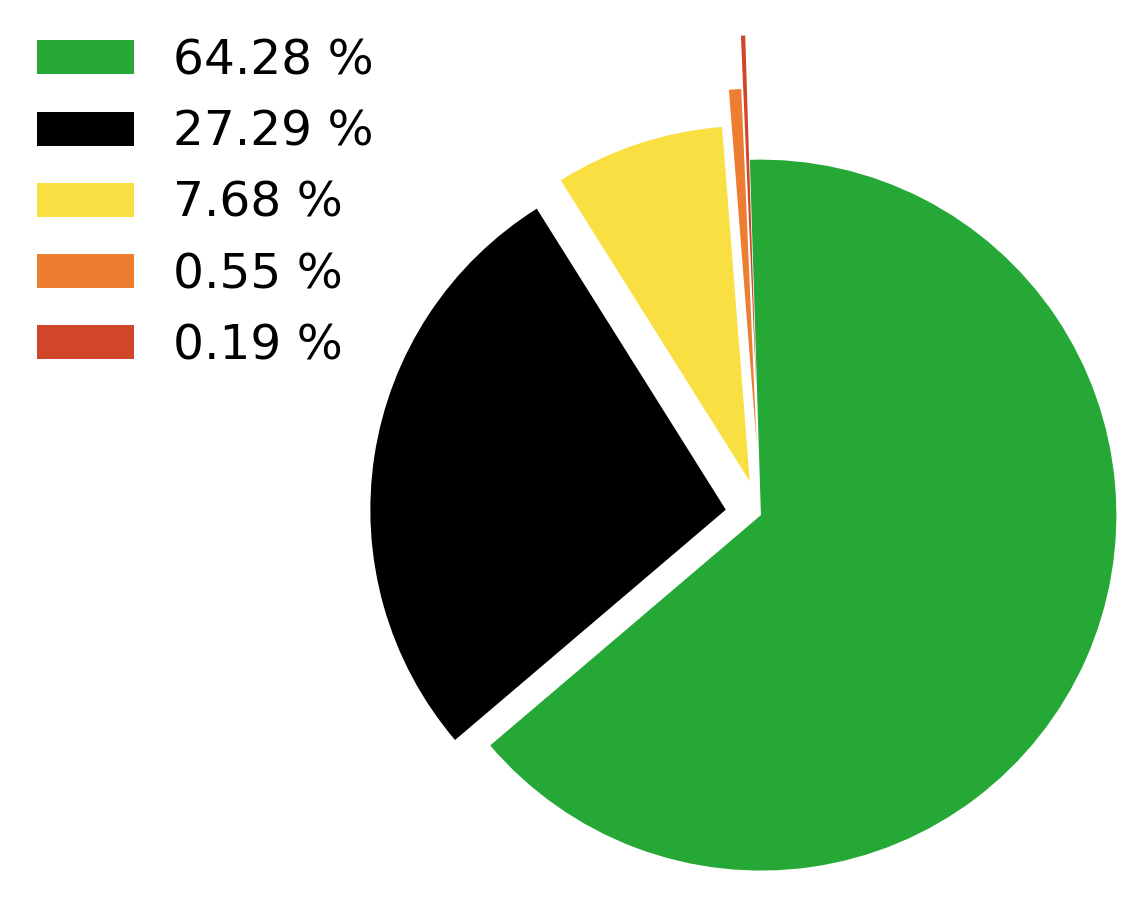}							\label{fig:graph17}}\hspace{\fill}
	\subfloat[Run\#4 - Model]{\includegraphics[width=0.48\columnwidth]{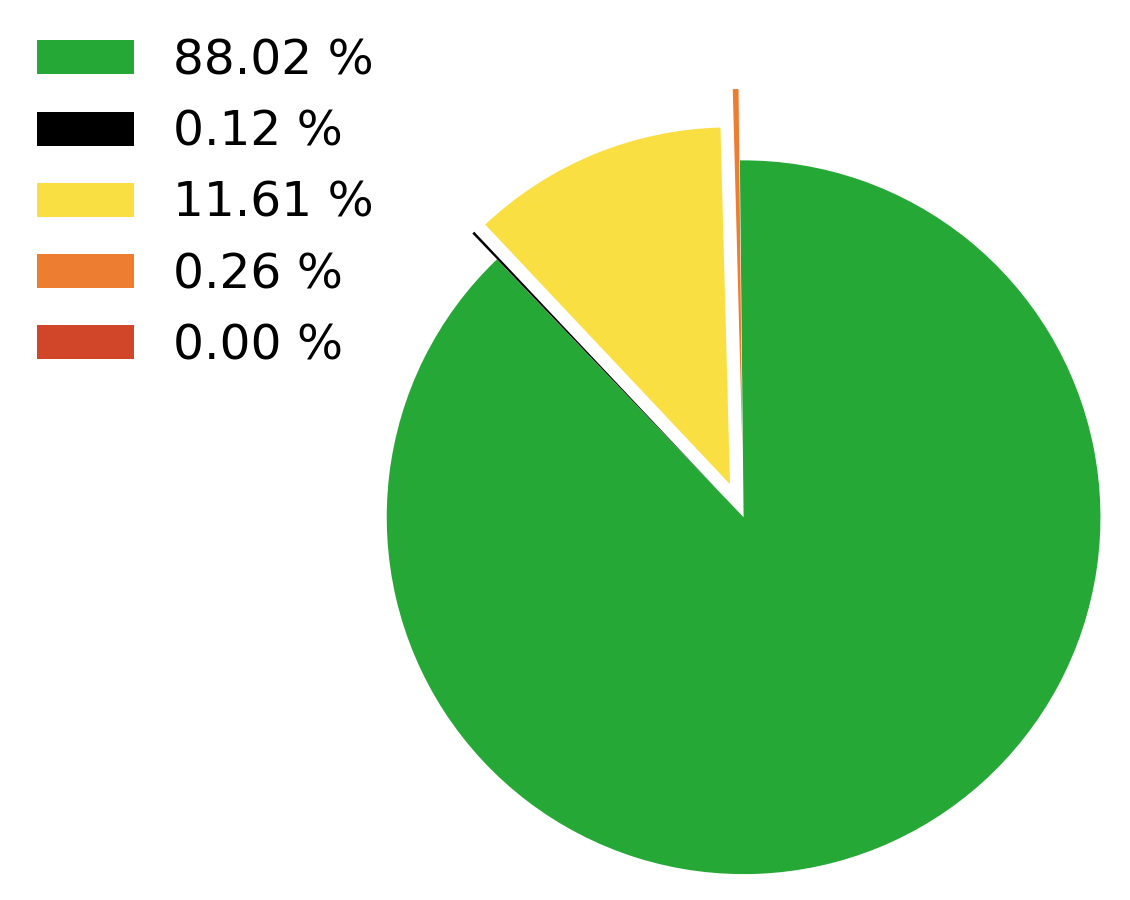}									
	\label{fig:graph18}}\\
	
	\subfloat[Run\#5 - Detector Only]{\includegraphics[width=0.48\columnwidth]{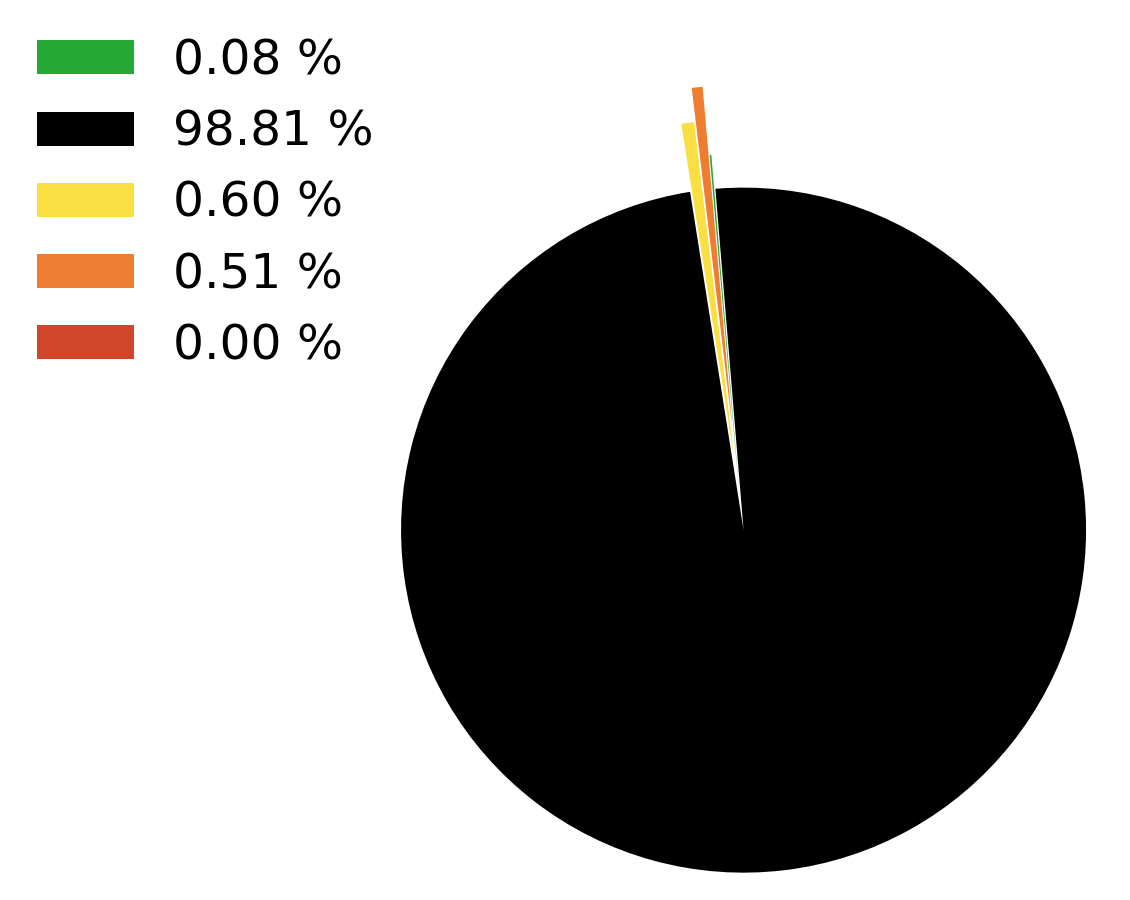}								\label{fig:graph13}}\hspace{\fill}
	\subfloat[Run\#5 - Model]{\includegraphics[width=0.48\columnwidth]{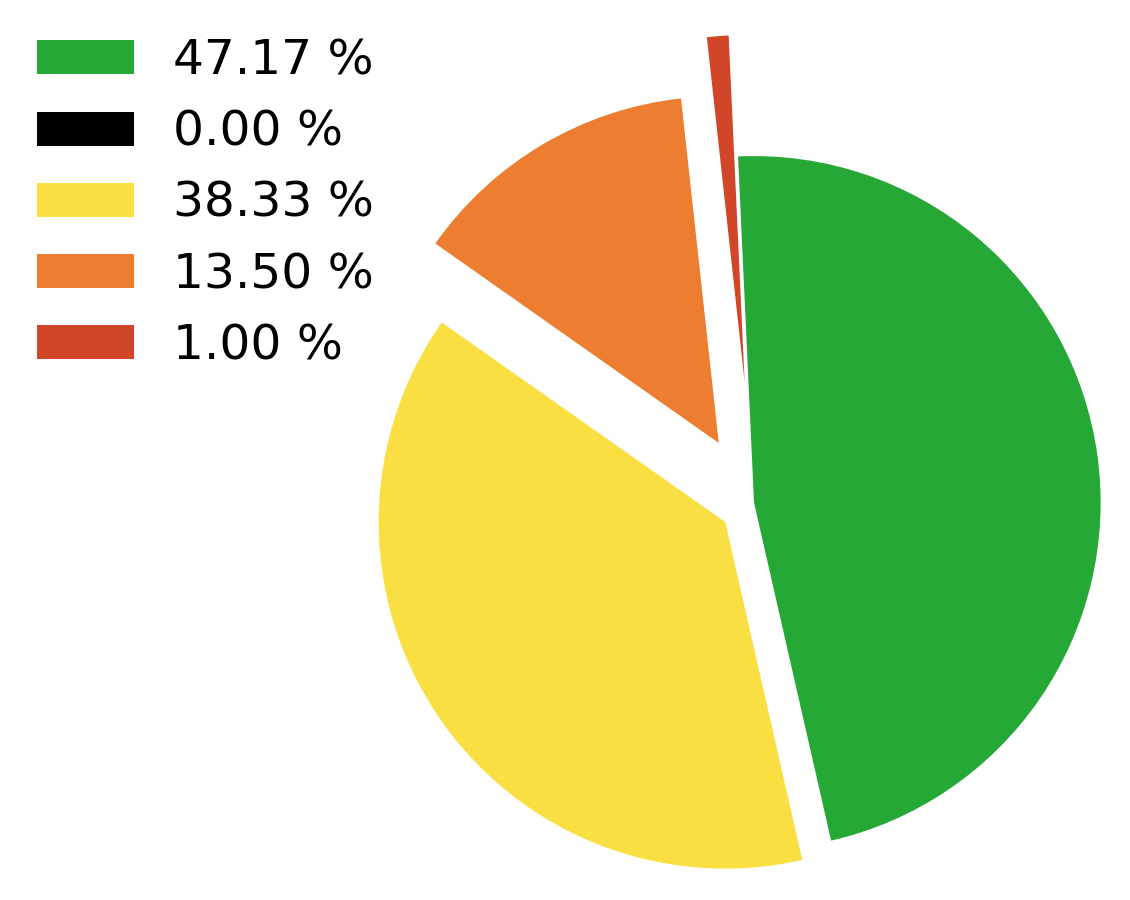}									
	\label{fig:graph14}}
	
  \caption{Comparison graphs between the localization accuracies using our proposal \wrt using both our proposal, in both monocular/stereo and line/clothoid tracking accordingly, and the MLD detector. The charts depict the results obtained on the Italian dataset in the A4 highway (4-lanes highway). A green color coding represents correct vehicle's lane localization, yellow represent one-lane range mismatches, orange 2-lane range mismatches and red 3-lane range mismatches. The black color represents the inability to assign a lane, due to missing information.}
  \label{fig:graph-all-italy}
\end{figure}

\begin{figure}[]
    \centering  	
  	\subfloat[Run\#6 - Detector Only] {\includegraphics[width=0.48\columnwidth]{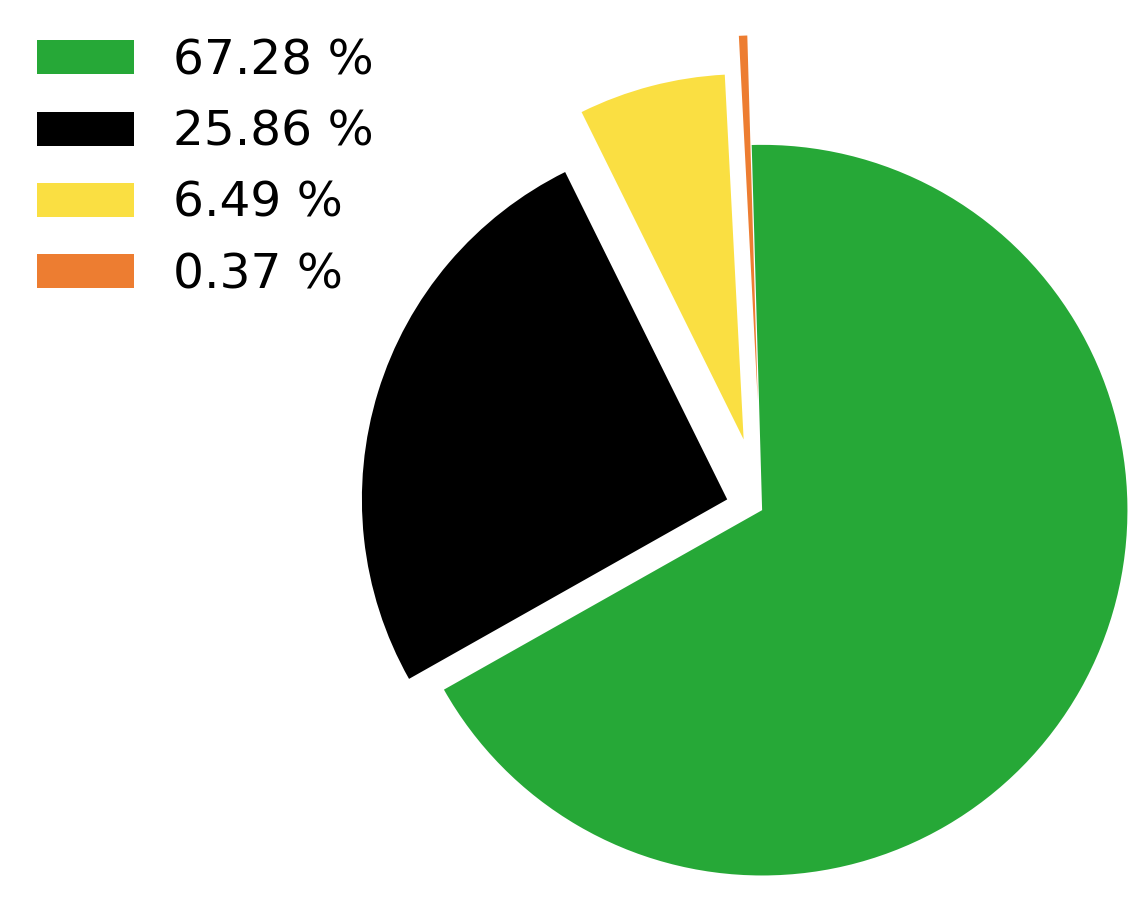}	\label{fig:graph5}}\hspace{\fill}
  	\subfloat[Run\#6 - Model] {\includegraphics[width=0.48\columnwidth]{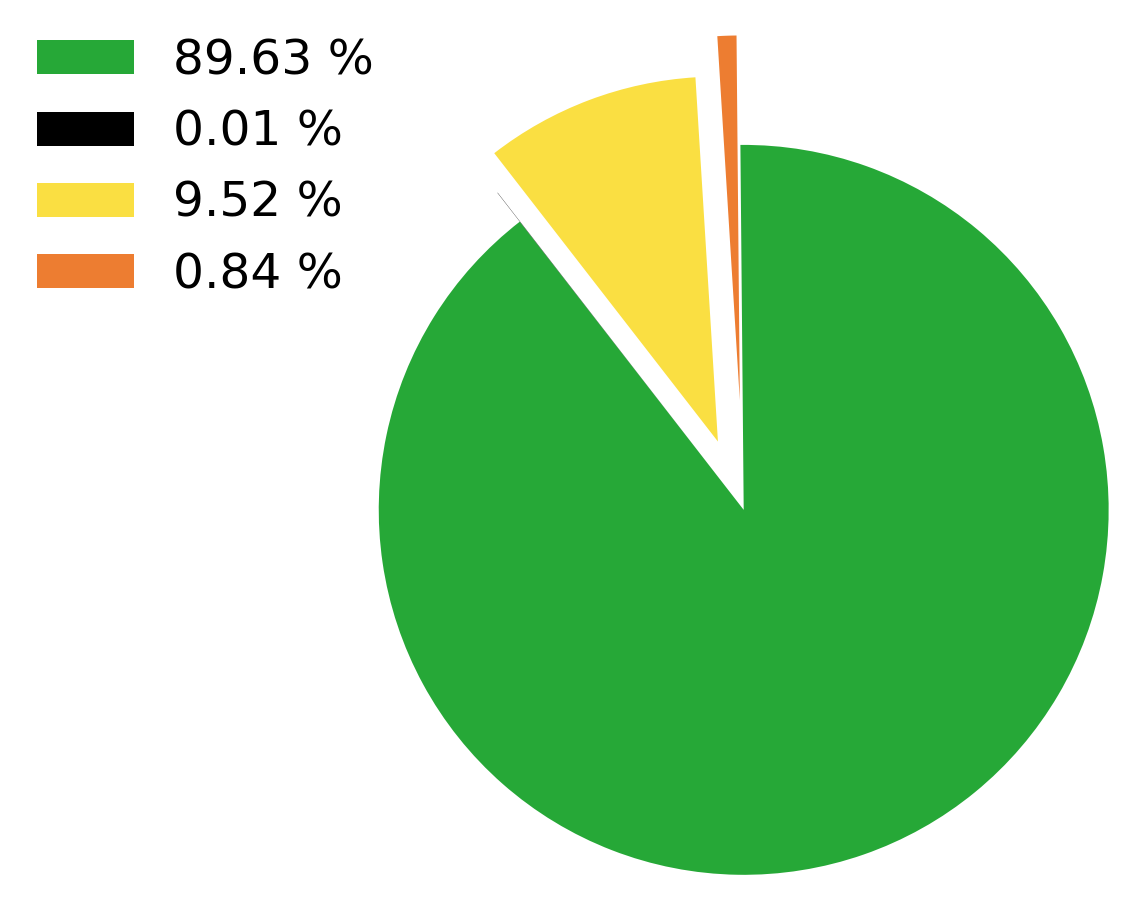}	\label{fig:graph6}}\\ 
  	
  	\subfloat[Run\#7 - Detector Only] {\includegraphics[width=0.48\columnwidth]{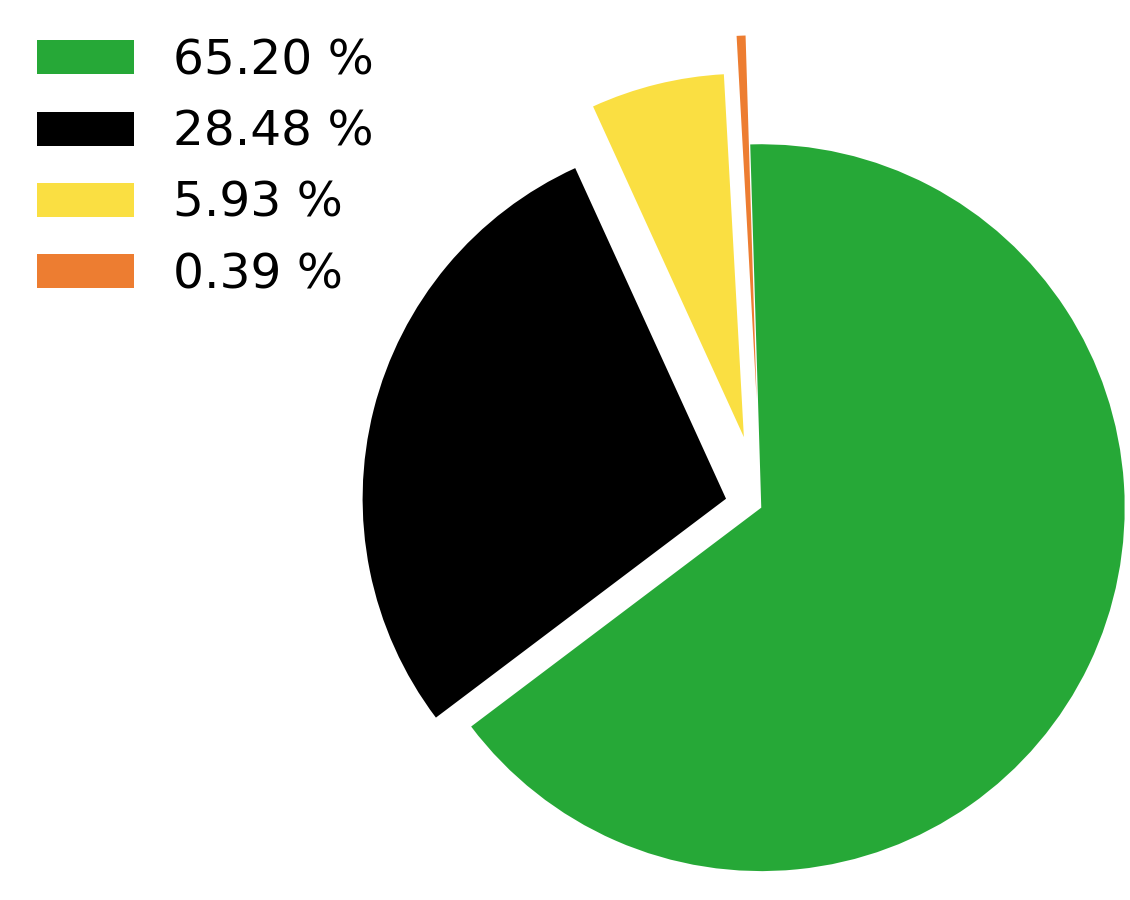}	\label{fig:graph9}}\hspace{\fill}
  	\subfloat[Run\#7 - Model] {\includegraphics[width=0.48\columnwidth]{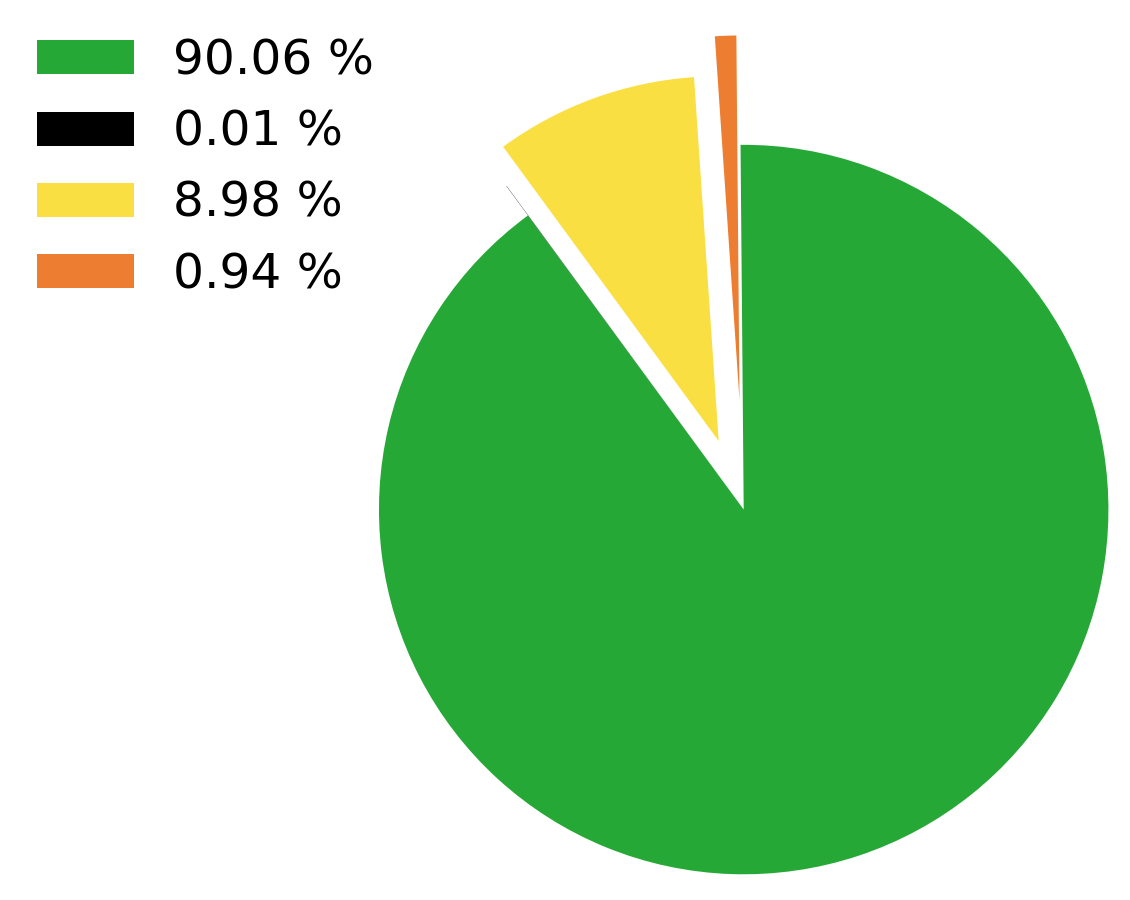}	\label{fig:graph10}}\\
  	
  	\subfloat[Run\#8 - Detector Only] {\includegraphics[width=0.48\columnwidth]{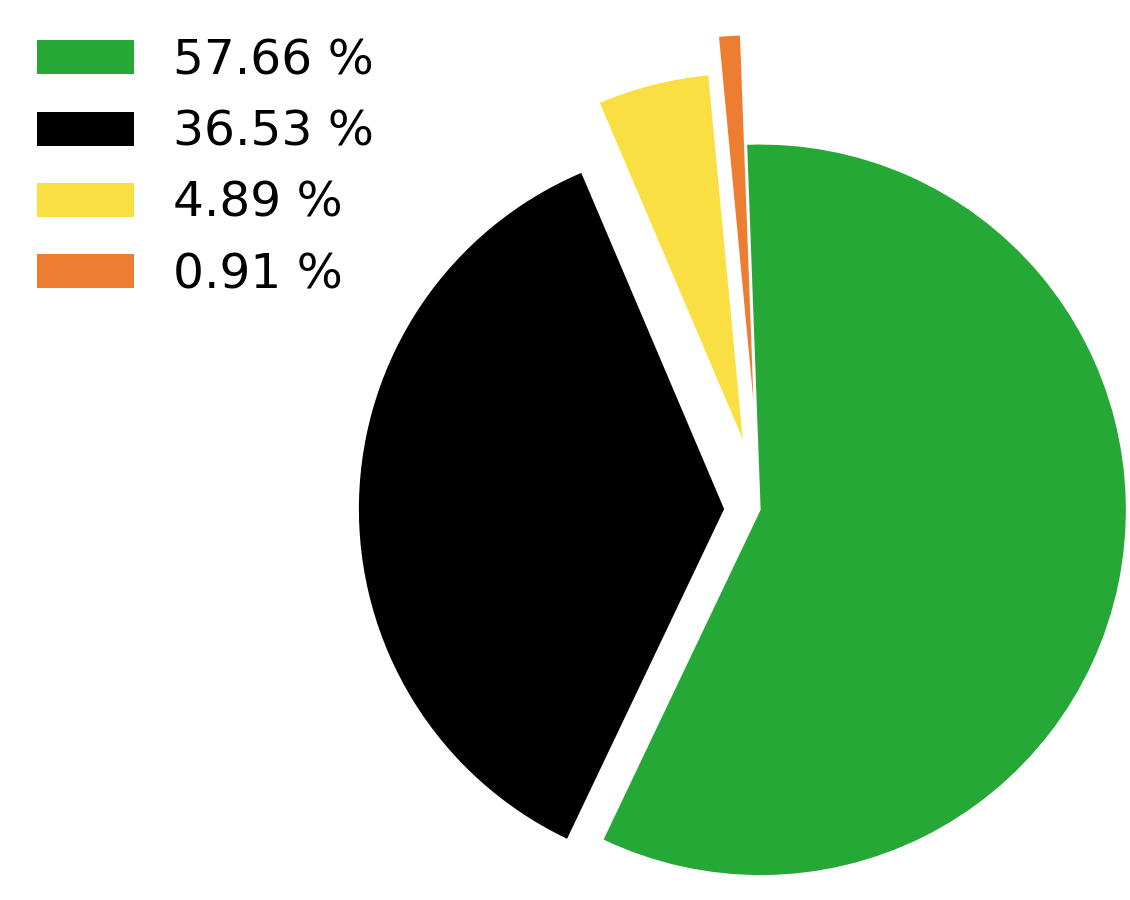}	\label{fig:graph7}}\hspace{\fill}
  	\subfloat[Run\#8 - Model] {\includegraphics[width=0.48\columnwidth]{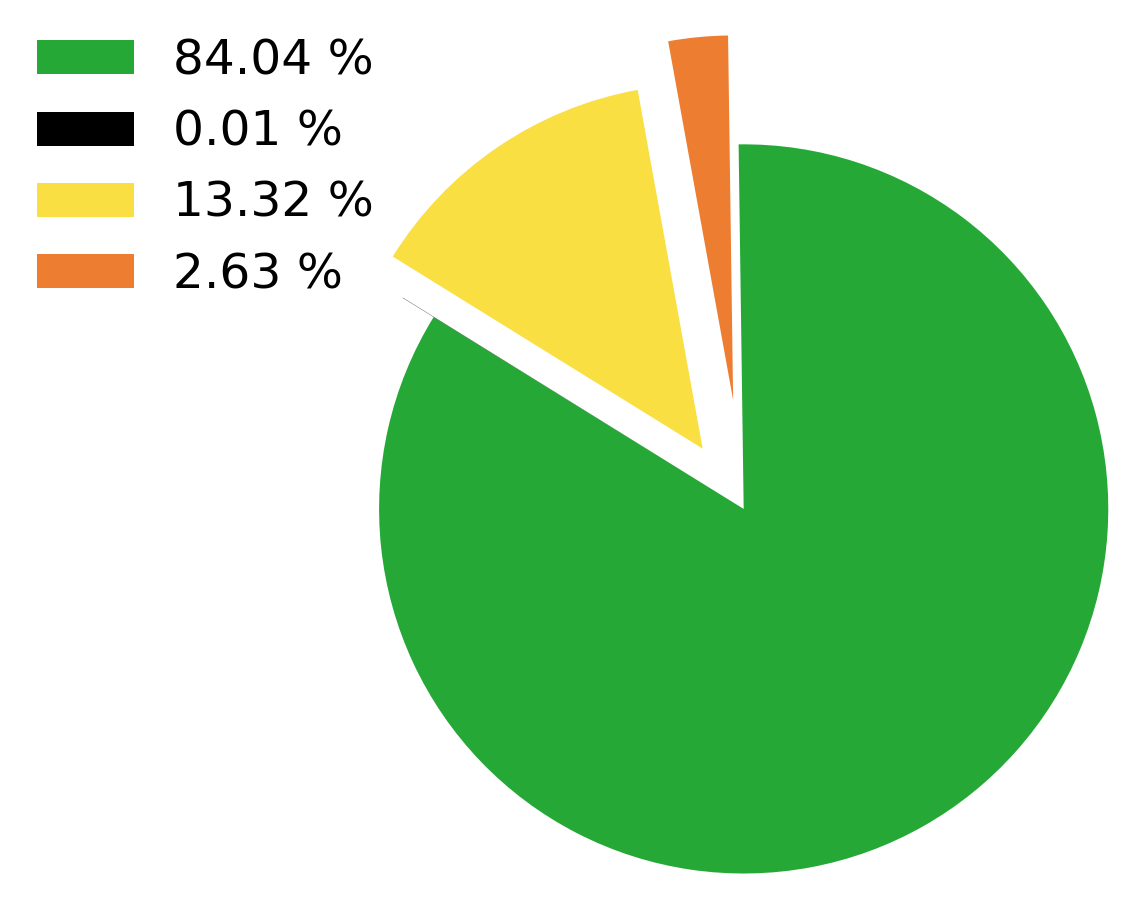}	\label{fig:graph8}}\\
         
	\subfloat[Run\#9 - Detector Only] {\includegraphics[width=0.48\columnwidth]{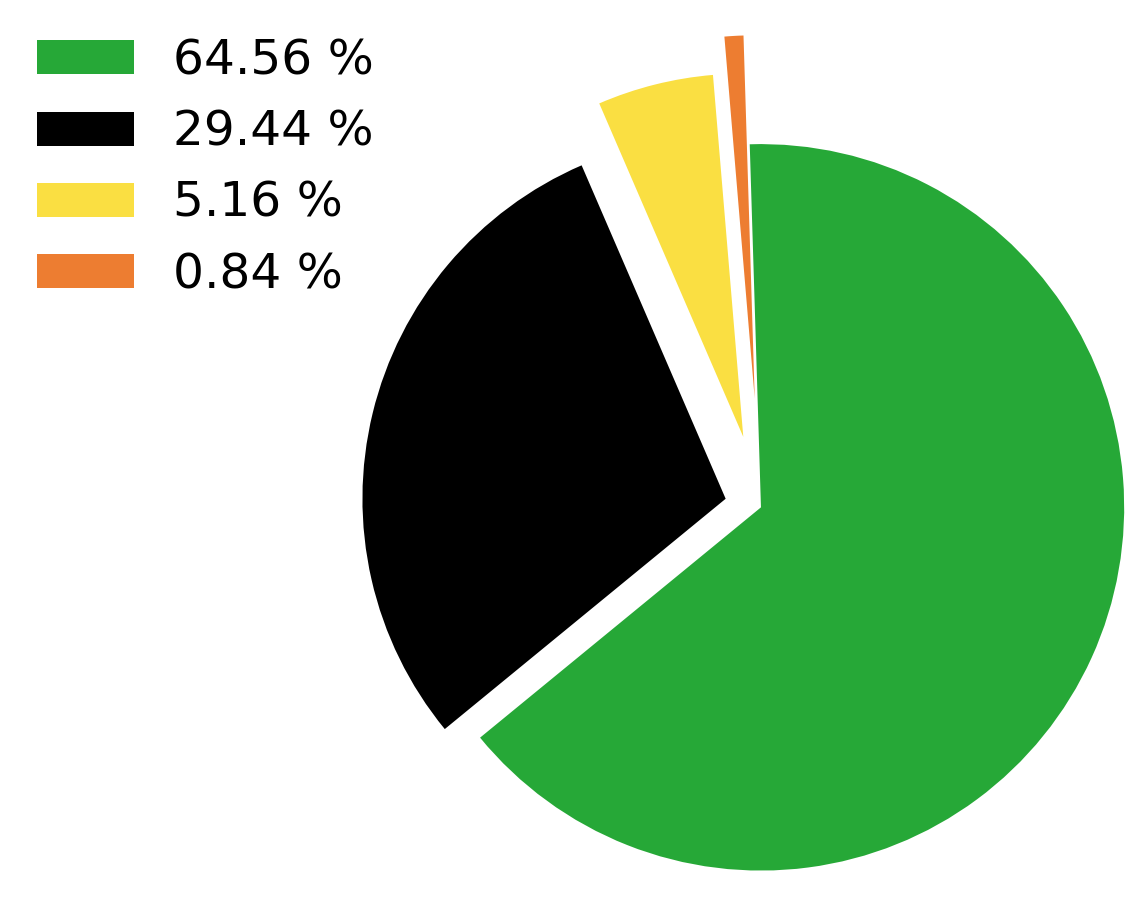}		\label{fig:graph11}}\hspace{\fill}
	\subfloat[Run\#9 - Model] {\includegraphics[width=0.48\columnwidth]{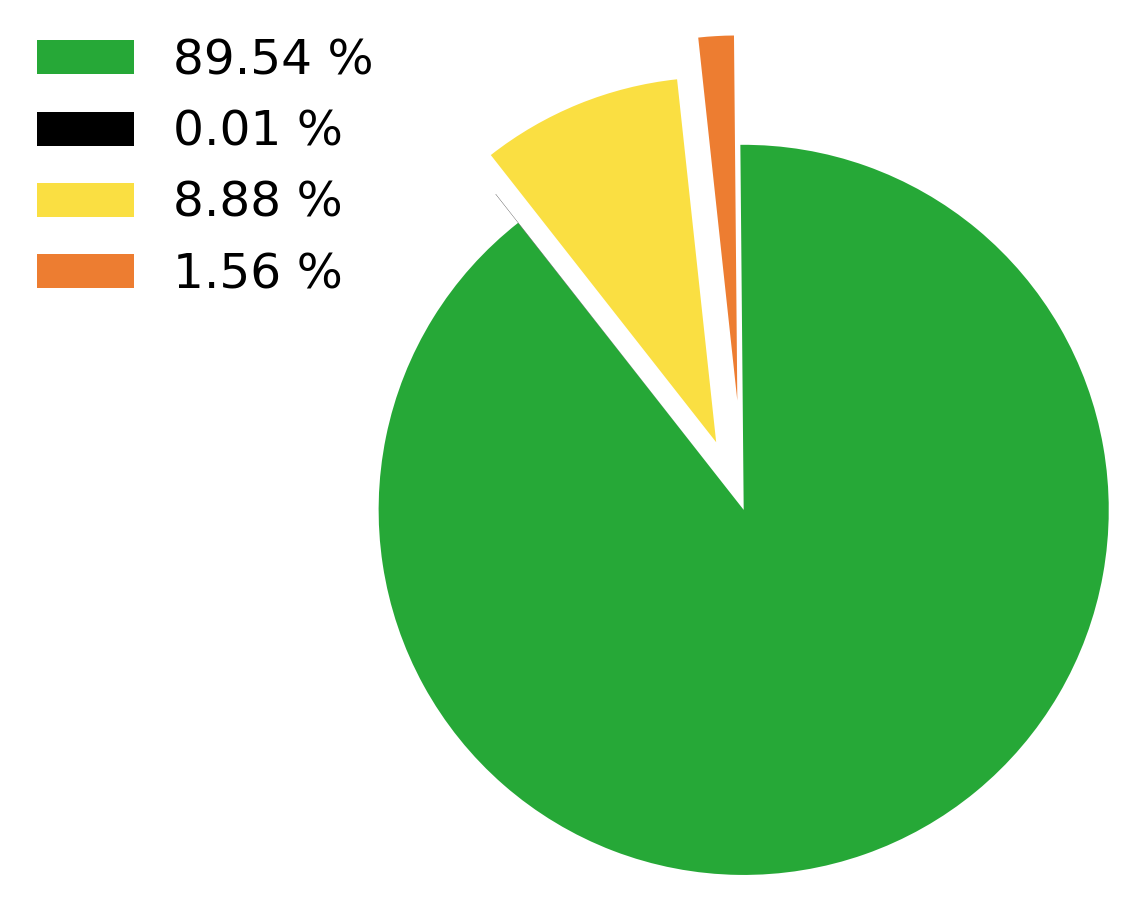}				\label{fig:graph12}}\\ 
    
    \subfloat[Run\#10 - Detector Only] {\includegraphics[width=0.48\columnwidth]{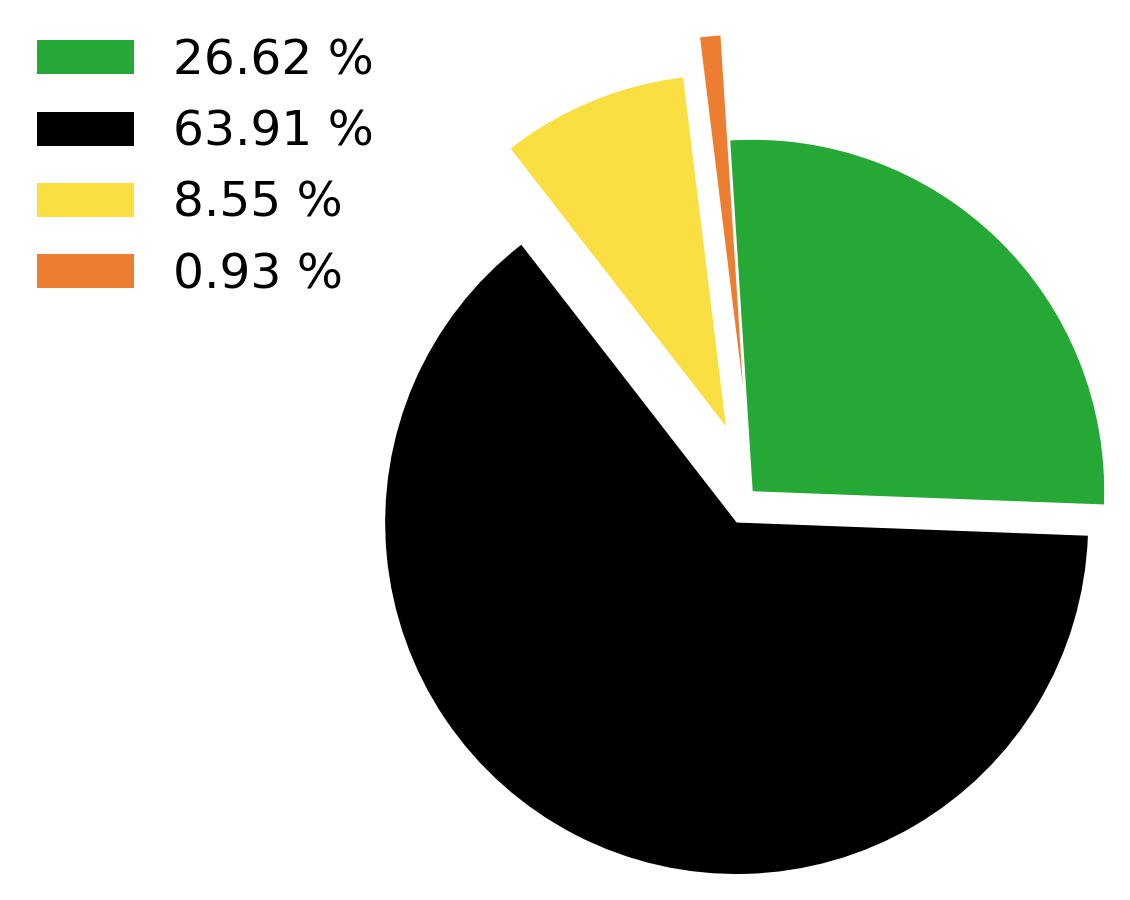}								\label{fig:graph19}}\hspace{\fill}
    \subfloat[Run\#10 - Model]{\includegraphics[width=0.48\columnwidth]{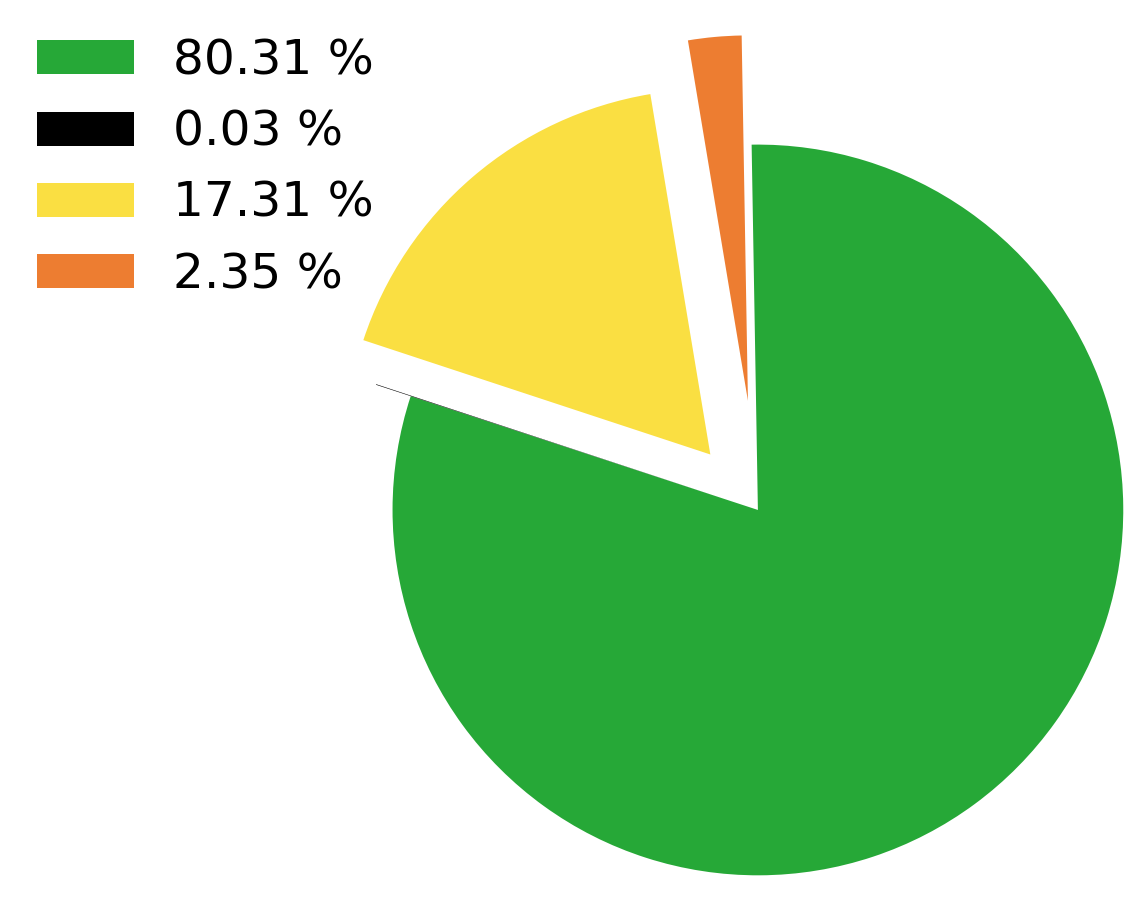}									
    \label{fig:graph20}}

  \caption{As in \Cref{fig:graph-all-italy} except it refers to the Spanish dataset, taken in the A2 highway (3-lanes highway).}
  \label{fig:graph-all-spain}
\end{figure}

\Cref{fig:lanetransition01} shows a short area of the A4 highway together with qualitative results of the algorithm, while in ~\Cref{fig:graph1,fig:graph2,fig:graph3,fig:graph4,fig:graph5,fig:graph6,fig:graph7,fig:graph8,fig:graph9,fig:graph10,fig:graph11,fig:graph12,fig:graph13,fig:graph14,fig:graph15,fig:graph16,fig:graph17,fig:graph18,fig:graph19,fig:graph20}
we report the dispersion over the ego-lane detection, considering all the frames of all experiments. Here the line detectors alone appear clearly unable to correctly detect the ego-lane, mostly because of missing detections due to clutter or illumination issues. In the figures this is depicted in black. A further confirmation of this can be found in Figure~\ref{fig:lanetransition01} middle: the detector only results are extremely noisy, resulting in an unreliable ego-lane determination. For instance, the detector is completely missing the final transition from $Lane_1$ to $Lane_2$, leaving the vehicle without almost any ego-lane localization clue. On the other hand, the filtering effect of the HMM is clearly shown in the same Figure top. The proposed model correctly identified the lane transitions even without a complete set of line measurements. Promising results are summarized in \Cref{fig:graph-all-italy,fig:graph-all-spain}. Our model outperformed the basic detector in all tests.

Concerning the confusion matrices \Cref{tab:run01model,tab:run01detector,tab:run02model,tab:run02detector,tab:run03model,tab:run03detector,tab:run04model,tab:run04detector,tab:run05model,tab:run05detector,tab:run06model,tab:run06detector,tab:run07model,tab:run07detector,tab:run08model,tab:run08detector,tab:run09model,tab:run09detector,tab:run10model,tab:run10detector}, it is worth noting that, for the dataset recorded in Spain, both all the algorithms and configuration settings achieve a better performance \wrt Italy. This is most likely related to the better view of the whole road in front of the vehicle, which contains 3 lanes instead of 4. We can also conclude that the localization results achieved by using our line detector are consistently higher, although this comparison might be unfair for the Italy A4 highway dataset, because of the limitations MLD algorithm (maximum 4 lines). However, our experiments prove the effectiveness of our proposal, as depicted in Figures \ref{fig:graph13} / \ref{fig:graph14}, and \ref{fig:graph19} / \ref{fig:graph20}.

\begin{figure}   
	\centering        	
	\subfloat[Run\#3 (No BV) - Detector Only] {\includegraphics[width=0.48\columnwidth]{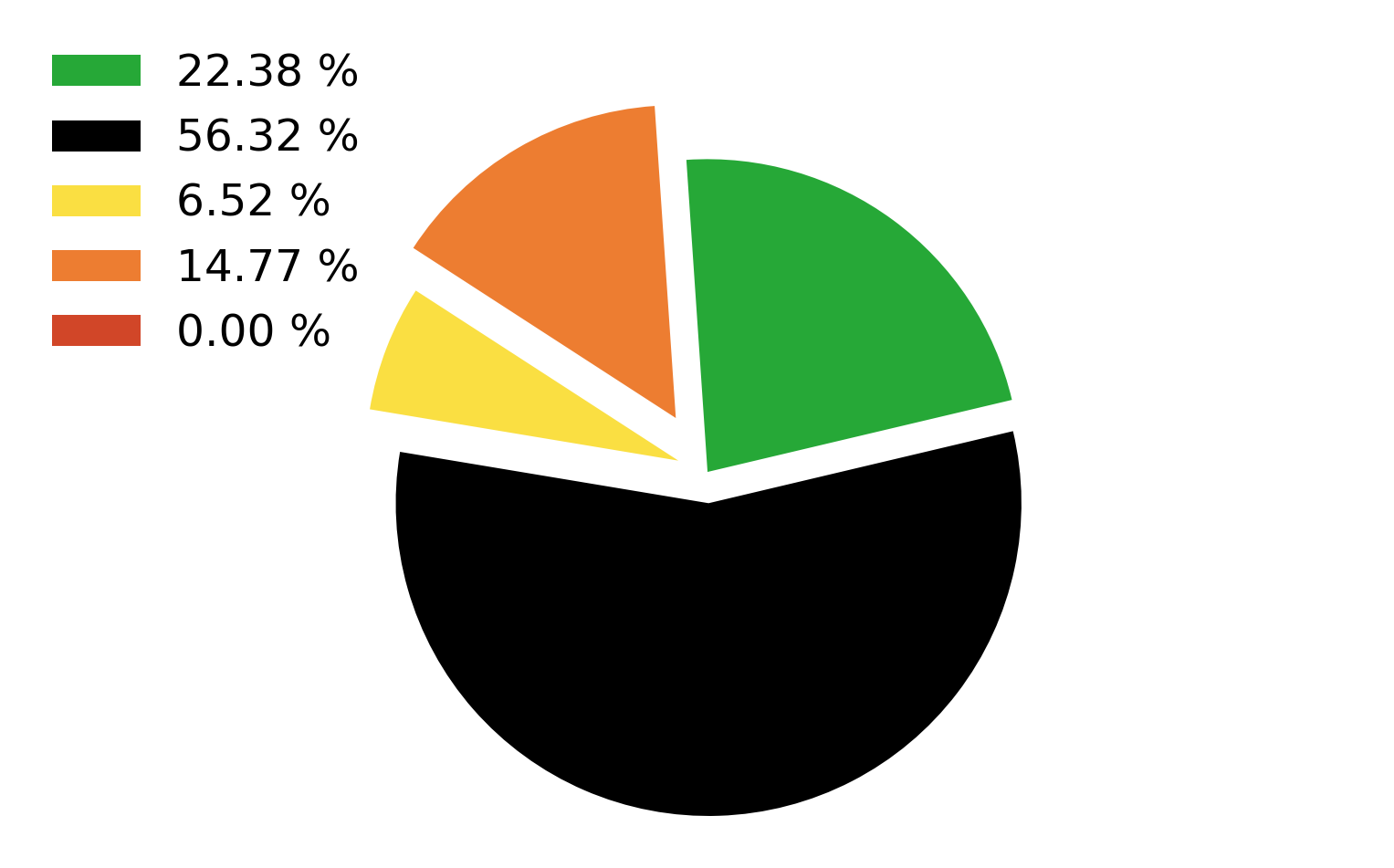}			\label{fig:graph15nobv}}\hspace{\fill}
	\subfloat[Run\#3 (No BV) - Model] {\includegraphics[width=0.48\columnwidth]{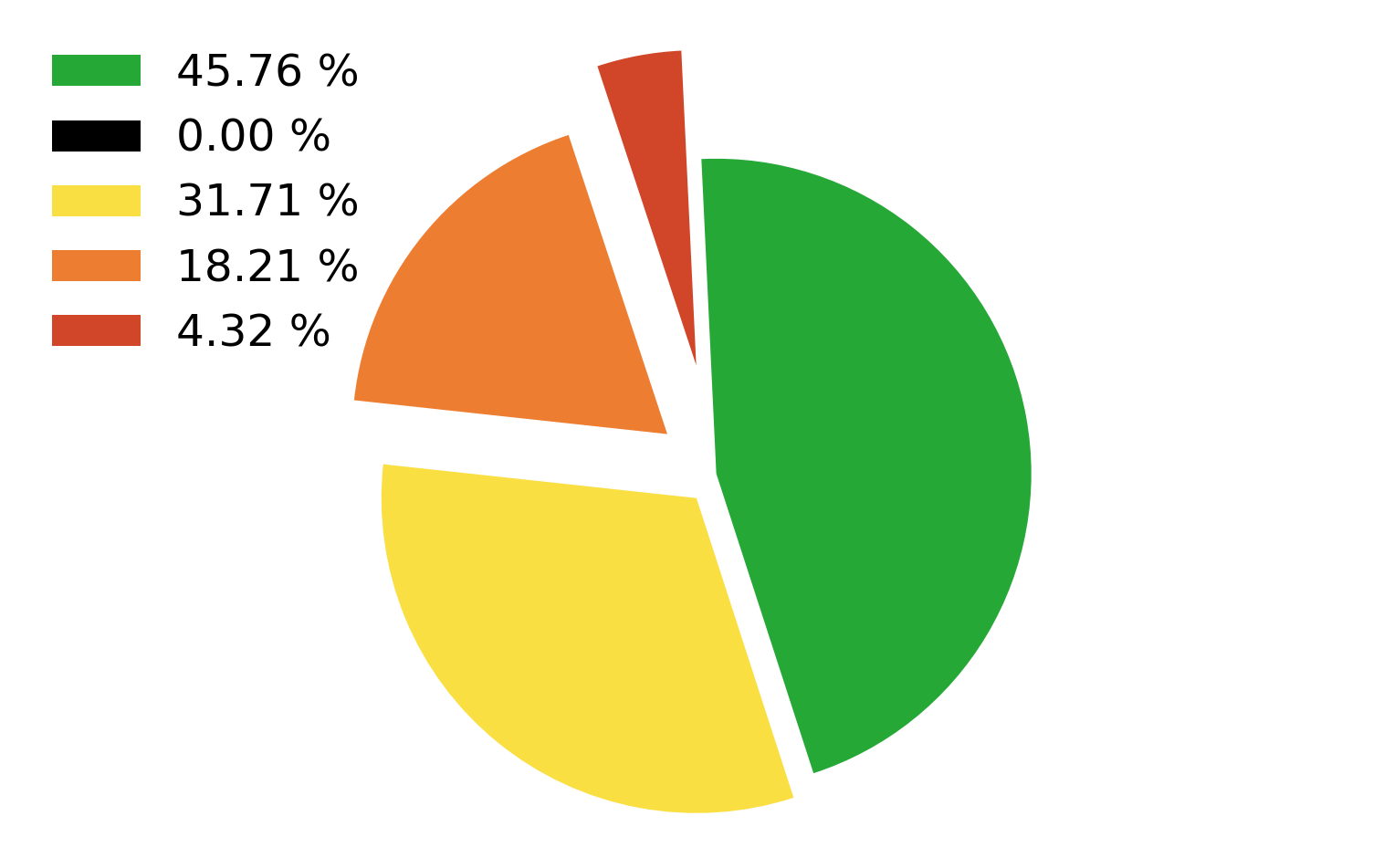}					\label{fig:graph16nobv}}
  \caption{Comparison of Run\#3 between the localization accuracies without the BV (see \Cref{fig:graph15,fig:graph16}).}
  \label{fig:comparisonBV-run3}
\end{figure}

From the result of the experiments we can observe that:
\begin{itemize}
\item using a line detector that is capable to classify the road line as dashed or continuous results in a better performance, see in \Cref{fig:comparisonBV-run3} an example of how the performance decreases when not using the BV.
\item in difficult situations that hampers the output of line detection and tracking, our approach is able to provide a valuable performance boost.
\end{itemize}

\section*{Conclusions}\label{sec:conclusions}
We presented an ego-lane estimation algorithm aimed at enhancing the accuracy of the vehicle localization at the lane level, in highway-like scenarios.
Differently from other works, we proposed a method designed to cooperate with an existing line detector and tracker.
With respect to the existing ego-lane estimation literature, our algorithm achieves good localization even when fed with noisy and/or occasionally missing data, \ie the typical output of a real, and therefore faulty, line detector and tracker.
We exploited an HMM-based scheme to take advantage of real road line observations in a probabilistic fashion.
The proposed algorithm improves the localization robustness in conditions where lane markings are missing or are hidden by traffic clutter and/or by lightening issues, \ie realistic conditions.
We are currently preparing to collect and analyze the results of the proposed algorithm in urban scenarios.

\bibliographystyle{IEEEtran}
\bibliography{t-iv.si-itsc17.egolaneestim}

\end{document}